\documentclass{article} 
\usepackage{iclr2027_conference,times}

\usepackage{amsmath,amsfonts,bm}

\def\eqref#1{equation~\ref{#1}}

\def\1{\bm{1}}

\DeclareMathAlphabet{\mathsfit}{\encodingdefault}{\sfdefault}{m}{sl}
\SetMathAlphabet{\mathsfit}{bold}{\encodingdefault}{\sfdefault}{bx}{n}

\usepackage{hyperref}
\usepackage{url}

\usepackage{xcolor}
\usepackage{amsmath}
\usepackage{enumitem}
\usepackage{subfigure}
\usepackage{stackengine}
\usepackage{graphicx}
\usepackage{epstopdf}
\usepackage{amssymb}
\usepackage{float}
\usepackage{booktabs}
\usepackage{amsthm}
\usepackage[english]{babel}
\usepackage{xpatch} 
\usepackage{diagbox}
\usepackage{algorithm}
\usepackage{algpseudocode}

\title{
Extending Rademacher Complexity-Based Generalization Theory to Domain Generalization
}

\author{
Xi Wang \& Liang Bai\thanks{Corresponding author} \\
Institute of Intelligent Information Processing \\
Shanxi University \\
Taiyuan, 030006, China \\
\texttt{\{wangxi7,bailiang\}@sxu.edu.cn}
\And
Xian Yang \\
Alliance Manchester Business School \\
University of Manchester \\
Manchester, M13 9PL, UK \\
\texttt{xian.yang@manchester.ac.uk}
\AND
Richard Yi Da Xu \\
Department of Mathematics \\
Hong Kong Baptist University \\
Hong Kong SAR, China \\
\texttt{xuyida@hkbu.edu.hk}
}

\newtheorem{Theorem}{Theorem}

\newtheorem{Corollary}{Corollary}
\newtheorem{Definition}{Definition}
\newtheorem{Assumption}{Assumption}
\newtheorem{Remark}{Remark}

\usepackage{tabu}
\usepackage{makecell}

\iclrfinalcopy 
\begin{document}


\maketitle
\fancyhead{}            
\ClearShipoutPicture    
\begin{abstract}
Domain generalization (DG) aims to train models on multiple source domains with distinct distributions to ensure robust performance on unseen target domains. In doing so, it elevates the concept of generalization from the traditional sample level to the distribution level. Existing theoretical frameworks introduce a domain space endowed with a meta-distribution to extend classical generalization analysis to this new setting. However, these approaches typically treat the domain space as devoid of geometry. In reality, the domain space is an inherently infinite-dimensional, nonlinear abstract space composed of probability distributions, and its complex geometric properties fundamentally determine generalization behavior. Consequently, existing geometry-free theories of domain generalization have a fundamental limitation. They lack a formal characterization of what distinguishes distribution-level generalization from sample-level generalization, leaving the inherent difficulty of generalizing across domains largely unexplained. To bridge this gap, we formalize the domain space as the Wasserstein space over the representation space. Building on the geometric properties of this space (specifically the $\mathrm{CD}(K,\infty)$ curvature theory) we derive a novel, geometry-dependent DG error bound. By introducing an additional curvature-dependent confidence term, this bound formally disentangles the two levels of generalization effects and reveals the essential nature that domain generalization is affected by both the invariance and the curvature of the representation space. Finally, we experimentally validate the effect of curvature on domain generalization.
\end{abstract}

\section{Introduction}\label{intro}
Domain generalization addresses a fundamental challenge in machine learning: enabling models to generalize effectively to unseen environments in the presence of distribution shift. In this setting, models are trained solely on data from multiple source domains, each with its own distinct distribution. During training, the model cannot access any samples or distributional information from the target domains. However, it must still generalize to arbitrary unseen targets. This stringent scenario raises a central theoretical conundrum: how can one bound the risk of a model on unknown target distributions without ever observing any target information?

To build intuition, it is instructive to revisit classical generalization theory. Generalization \citep{vapnik71} is arguably the most fundamental problem in machine learning, as it quantifies a model's ability to perform well on unseen data after training on a finite set of examples.
Among various approaches in statistical learning theory \citep{valiant1984theory0, Bousquet75, Huan74, Bartlett95, haussler111, dudley112}, the Rademacher complexity-based generalization theory \citep{bartlett72, bartlett73} has become one of the most influential frameworks for characterizing generalization performance.
It leverages measure concentration inequalities (e.g., McDiarmid's inequality \citep{McDiarmid78}) to explicitly capture the impact of sample randomness on generalization and derives a generalization error bound based on the model’s Rademacher complexity \citep{bartlett72, bartlett73}. The bound is controlled by two terms: the first is the Rademacher complexity term of the model, which characterizes the model's ability to fit the training samples; 
the other is a confidence term that quantifies the impact of sample randomness on generalization, arising directly from the application of measure concentration inequalities in the derivation of the bound.

Inspired by classical generalization theory, existing work formalizes domain generalization by introducing a domain space (the space of domain distributions) equipped with a meta-distribution \citep{Li90}. Under this view, each domain is a point in the domain space; source domains act as training samples, and target domains act as unseen test points. Domain generalization thus becomes a two-level generalization problem: the model must generalize across samples within each domain, and simultaneously across domains in the abstract domain space. Accordingly, the domain generalization error is modeled as an outer expectation over the meta-distribution of the inner expectations over the per-domain sample distributions, and the domain generalization problem is analyzed by bounding this quantity. This analogy elegantly lifts the idea from the sample level to the domain level. However, it has a critical flaw: current theory treats the domain space as a geometry-free space, essentially no different from the sample space, while ignoring the fact that, as an infinite-dimensional, nonlinear space of probability distributions, it possesses a complex geometric structure. These geometric properties fundamentally alter how cross-domain generalization works. Consequently, the resulting domain generalization error bounds cannot cleanly separate sample-level and domain-level generalization effects, and thus fail to reveal the fundamental distinction between generalization at these two levels. There is therefore an urgent need for a rigorous foundation of the domain space that incorporates a well‑defined geometric structure.

To this end, this paper proposes that the domain space be conceptualized as a Wasserstein space. This choice is not merely natural but also principled, for three reasons. First, there is a bijective correspondence between domains and their associated probability distributions, and the space of all probability distributions over the sample space is precisely a Wasserstein space. Second, the Wasserstein distance directly measures the optimal transport cost between distributions, aligning perfectly with the notion of distribution shift that domain generalization aims to handle. Third, and most importantly for theory, the Wasserstein space possesses rich geometric properties that have been extensively studied in the existing literature on Wasserstein geometry \citep{Lott91}. These results can be readily adapted to the study of domain generalization. This provides a geometrically grounded approach to characterizing generalization behavior in the domain space. Following the paradigm of classical generalization theory, the key question then becomes how to define the concepts of generalization in this abstract space that has been endowed with geometric structure.
To this end, we use the Wasserstein space to define in detail various concepts of the domain generalization problem, including distribution-level Rademacher complexity, invariance of the representation space, and so on. On this basis, we combine the well-known \(\mathrm{CD}(K,\infty)\) curvature theory with the McDiarmid concentration inequality. The resulting measure concentration inequality not only reflects how well the representation space captures cross-distribution invariance, but also depends on its lower Ricci curvature bound. Building on these theoretical foundations, we derive a domain generalization error upper bound that explicitly depends on curvature. Our bound shows that the model's ability to generalize to unseen domains is jointly determined by the curvature and the invariance of the representation space. 

The primary contributions of this paper are listed as follows:
\begin{itemize}
    \item We formalize the domain space as a Wasserstein space, and define various concepts for domain generalization, including distribution-level Rademacher complexity and invariance of the representation space.
    \item We are the first to introduce Wasserstein geometry into domain generalization. By modifying the McDiarmid-type concentration inequality using the \(\mathrm{CD}(K,\infty)\) curvature theory, we lay the mathematical foundation for generalization analysis on the Wasserstein space.
    \item  We derive a domain generalization error bound that isolates the distribution-level confidence term, revealing that domain generalization is affected by the curvature of the representation space and the degree to which it captures cross-distribution invariance.
\end{itemize}

\section{Preliminaries}
\label{pre}
In this section, we introduce the main foundations of this paper, including key definitions, notations, formulas and some necessary assumptions.

\subsection{Rademacher Complexity-based Statistical Learning Theory}
\label{subsec_claG}
We first introduce classical Rademacher complexity-based statistical learning theory, beginning with the generalization error—the key quantity that this theory aims to bound.

We consider a hypothesis class $\mathcal{H}$ defined over the input space $\mathbb{X} \subset \mathbb{R}^{d}$, where $d$ denotes the dimension. Let $\hat{\mathbb{Y}} \subset [0,1]$ be the output space and $\mathbb{Y} \subset [0,1]$ the labeling space. Let $P$ be a probability distribution on sample space $\mathbb{X} \times \mathbb{Y}$, and let $S=\{(x_i, y_i)\}_{i=1}^n$ be $n$ i.i.d. samples drawn from $P$.

For any hypothesis $h \in \mathcal{H}$, the expected risk on distribution $P$ is defined as:
\begin{equation}\label{risk1}
    \mathcal{E}_P(h) = E_{(x,y) \sim P}[\ell(h(x), y)],
\end{equation}
while the empirical risk on samples $S = \{(x_i,y_i)\}_{i=1}^n \subset \mathbb{X}$ is:
\begin{equation}
    \hat{\mathcal{E}}_S(h) = \frac{1}{n} \sum_{i=1}^n \ell(h(x_i), y_i),
\end{equation}
where the loss function $\ell$ is $q$-Lipschitz continuous with respect to its first argument. 
The generalization error is defined as: 
\begin{equation}\label{Err0}
    \mathrm{Err}(h) = \mathcal{E}_P(h) - \hat{\mathcal{E}}_S(h).
\end{equation}
\textbf{Remark.} In this paper, the following notations are considered equivalent: $E_{x \sim P}[y]$, $E_{x}[y]$ and $E_{P}[y]$.

\noindent Then, the empirical Rademacher complexity of $\mathcal{H}$ conditioned $n$ samples is:
\begin{equation}
\mathcal{\hat{R}}_n(\mathcal{H}) = E_\sigma \left[ \sup_{h \in \mathcal{H}} \frac{1}{n} \sum_{i=1}^n \sigma_i h(x_i) \Big| x_1, \dots, x_n \right],
\end{equation}
where $\sigma_i \sim \text{Unif}\{\pm 1\}$ are Rademacher variables. The (population) Rademacher complexity is defined as the expectation of $\mathcal{\hat{R}}_n(\mathcal{H})$ over the random draws of the samples:
\begin{equation}
\mathcal{R}_n(\mathcal{H}) = E_{x_1, \dots, x_n} \left[ \mathcal{\hat{R}}_n(\mathcal{H}) \right].
\end{equation}

This measure characterizes the maximum extent to which a hypothesis class can fit purely random labels on the given samples. A larger Rademacher complexity indicates that the hypothesis class is rich enough to capture spurious patterns, thereby signaling a higher propensity for overfitting and worse generalization.

The following theorem provides an upper bound on the generalization error and reveals the relationship between model complexity and generalization.
\begin{Theorem}[Generalization Error Bound \citep{bartlett72}]\label{thm_CG}
For any $\delta \in (0,1)$ and $h \in \mathcal{H}$, the following inequality holds with a confidence level of at least $1-\delta$:
\begin{equation}
E[h(x)] \leq \frac{1}{n}\sum_{i=1}^nh(x_i) + 2\hat{\mathcal{R}}_n(\mathcal{H}) + \mathcal{O}\Big(\sqrt{\frac{\ln(1/\delta)}{n}}\Big),
\end{equation}
More generally, for the generalization error:
\begin{equation}\label{eqn7}
    \sup_{h \in \mathcal{H}} \mathrm{Err}(h) \leq \underbrace{2q\hat{\mathcal{R}}_n(\mathcal{H})}_{\text{Complexity term}} + \underbrace{\mathcal{O}\Big(\sqrt{\frac{\ln(1/\delta)}{n}}\Big)}_{\text{Confidence term}}.
\end{equation}
where the $q$ is the Lipschitz coefficient of loss function $\ell$ with respect to its first argument.
\end{Theorem}
The second term on the R.H.S. of Eqn.~\ref{eqn7} is the high‑probability confidence term for the Rademacher complexity, which is grounded in the following probabilistic foundation:

\noindent \textbf{Measure Concentration Foundation:}
McDiarmid's inequality \citep{McDiarmid78} is a fundamental probabilistic tool in statistical learning theory, and is given as follows:\\  
If a function $g: \mathbb{X}^n \to \mathbb{R}$ satisfies the bounded difference condition,
\begin{equation}
    \sup_{x_1,\dots,x_n, x_i^{\prime}} |g(x_1,\dots,x_i,\dots,x_n) - g(x_1,\dots,x_i^{\prime},\dots,x_n)| \leq c_i,
\end{equation}
then for any $t>0$ it satisfies the following exponential concentration:
\begin{equation}
    P\left( |g(x_1,\dots,x_n) - E_{P^n}[g(x_1,\dots,x_n)]| \geq t \right) \leq 2\exp\left(-\frac{2t^2}{\sum_{i=1}^n c_i^2}\right),
\end{equation}
where $x_1, \cdots, x_n, x_i^{\prime} \in \mathbb{X}$, $i \in [n]$.

The inequality in Theorem~\ref{thm_CG} gives a high‑probability upper bound for the generalization error, comprising a Rademacher complexity term that quantifies the model complexity on the finite training sample, along with a confidence term that provides a probabilistic guarantee for the complexity measure. Here, the confidence term arises from statistical laws, while the essence of generalization lies in the model complexity. A model with high complexity tends to overfit the training samples and thus loses its generalization capability, resulting in a larger generalization bound.

\subsection{Basic Notations and Definitions of Domain Generalization}
\label{subsec_Pre2}
We first introduce the notation used in domain generalization. Domain generalization (DG) considers the setting where models are trained on samples drawn from multiple source domains and are required to generalize to a broader range of target domains, including unseen domains, with each domain following a potentially different distribution. Let \(\mathrm{Enc}(\cdot):\mathbb{X}\to\mathbb{Z}\) be the encoder under consideration, and for any target domain distribution \(P\) on the original input space \(\mathbb{X}\), let \(D:=\mathrm{Enc}_{\#}(P)\) be its pushforward measure on the representation space \(\mathbb{Z}\). In the sequel, we represent a domain by this induced distribution \(D\) over \(\mathbb{Z}\), and assume that each domain is associated with a labeling function \(f:\mathbb{Z}\rightarrow\mathbb{Y}\) (following \cite{Albuquerque38, Zhao15, Redko93}). Let \(\mathbb{D}\) denote the set of all possible target domain distributions, and let \(\hat{\mathbb{D}}_m \subset \mathbb{D}\) be the set of \(m\) domains randomly drawn from \(\mathbb{D}\), referred to as the source domains. The \(i\)-th source domain is denoted by \((D_i, f_i)\); under the covariate shift assumption (Assumption~\ref{CSA}), we write it simply as \(D_i\). We denote the distribution over \(\mathbb{D}\) by \(\mathcal{P}\) and call it the meta-distribution. Furthermore, we assume that each source domain contains \(n\) samples: the \(j\)-th sample from the \(i\)-th source domain is denoted by \(x_{i,j}\), the sample set of the \(i\)-th source domain is denoted by \(S_i\), and \(S\) denotes the collection of all \(mn\) samples.

Under above concepts, the domain generalization error is defined as following:
    \begin{equation}
    \mathrm{Err}_D(h) = E_{D \sim \mathcal{P}}\mathcal{E}_D(h) - \frac{1}{m}\sum_{i=1}^m \hat{\mathcal{E}}_{S_{i}}(h),
\end{equation}
where $\mathcal{E}_D(h)$ denotes the risk obtained by replacing the label $y$ in Eqn.~\ref{risk1} with the labeling function.

Based on the above definitions, there has been work that used Rademacher complexity to investigate domain generalization and provided a domain generalization error bound. 
\begin{Theorem}[Domain Generalization Error Bound \citep{Li90}]\label{thm:DG2022}
Under Assumption~\ref{CSA}, and assume the loss function $\ell$ takes values in $[0,1]$. For any hypothesis class $\mathcal{H}$ and any domain $d \in \mathbb{D}$, with probability at least $1 - \delta$ over the random sampling of $S$, the following holds uniformly for all $h \in \mathcal{H}$:
\begin{equation}
   \mathrm{Err}_D(h) \leq \frac{2}{m} \sum_{i=1}^m \underbrace{\mathcal{\hat{R}}^i_m(\ell \circ \mathcal{H})}_{\text{Per-domain Complexity}} + 2 \underbrace{\mathcal{\hat{R}}_{mn}(\ell \circ \mathcal{H})}_{\text{Joint Complexity}} + 5 \underbrace{\sqrt{\frac{\ln(3/\delta)}{2mn}}}_{\text{Confidence term}}. 
\end{equation}
where the $\mathcal{\hat{R}}^i_m(\ell \circ \mathcal{H})$ denotes the empirical Rademacher complexity with respect to $S_m^i$ (the set formed by taking the $i$-th sample from each source domain), and $\mathcal{\hat{R}}_{mn}(\ell \circ \mathcal{H})$ denotes the empirical Rademacher complexity with respect to the training samples from all source domains.
\end{Theorem}
\begin{Remark}
    \citep{Li90} uses the conditional distributions $P(y|x)$ to represent labeling functions and also adopts the covariate shift assumption.
\end{Remark}
This result is an early attempt to understand domain generalization from the perspective of Rademacher complexity, revealing the additional impact of model complexity on cross-domain generalization. However, this analytical approach essentially treats the domain space $\mathbb{D}$ as a space devoid of geometric structure and indiscriminately transplants Rademacher-complexity tools together with probabilistic concentration tools onto the domain space, failing to capture the distinctive nature of distribution-level generalization. Therefore, in order to meaningfully distinguish the specificity of cross-distribution generalization from sample-level generalization, a deeper and more specialized theoretical framework is required to elucidate the fundamental mechanisms of domain generalization. In what follows, we abstract the domain space as a Wasserstein space (an infinite-dimensional nonlinear space of probability measures) and aim to exploit its distinctive geometric properties to establish a theoretical analysis of generalization on this space.

\section{Main Result}\label{MainCon} 
In this section, we develop a generalization analysis framework on the Wasserstein space. First, we provide the definition of domain generalization on this space, and define an invariance measure of the representation space as well as a Rademacher complexity representation of the model with respect to the source domain distributions. Then, using existing results in Wasserstein geometry, we establish a McDiarmid-type measure concentration inequality tailored to our setting, which connects its tail bound to the invariance of the representation space and the lower bound on Ricci curvature. Finally, based on the above analysis, we obtain the final domain generalization error bound.

\subsection{Definitions of Domain Generalization on Wasserstein Space}\label{subsec_DS-WS}
We first introduce an assumption and some formal definitions of domain generalization on the Wasserstein space, covering both the sample space and the target domain space.
\begin{Definition}[Sample Space]
\label{def:ssds}
For a representation space $\mathbb{Z}$, we define the sample space as a metric-measure space $(\mathbb{Z}, d, \nu)$, where $d$ and $\nu$ are the metric and reference measure on $\mathbb{Z}$.
\end{Definition}
For the definition of metric-measure space, as well as the contents of the Wasserstein space and Wasserstein geometry used in the following, see Appendix~\ref{Add-Notation} for details.

\begin{Assumption}
\label{ass:lp-bounded}
We assume that exists a representation space $\mathbb{Z}$ such that for any target domain, its induced distribution on $\mathbb{Z}$ is absolutely continuous with respect to the reference measure $\nu$, and the Radon–Nikodym derivative belongs to $L^2$ space, i.e. $D \ll \nu$ and $\frac{\rm{d}D}{\rm{d\nu}} \in L^2(\mathbb{Z})$.
\end{Assumption}
\begin{Remark}
    Since unrestricted domain generalization is infeasible, we constrain the target domain by requiring all domain distributions to be sufficiently close to the reference distribution $\nu$. This mirrors the reference-set idea in \cite{Albuquerque38}, \cite{Sicilia57}, and \cite{Ye44}, which also assume restricted target domains. In the following, we will further quantify the difficulty of domain generalization by defining the concept of the $C$-regular domain generalization problem.
\end{Remark}

\begin{Definition}[Domain Space]
\label{def:ssds2}
We define the space of all possible domains as the $2$-Wasserstein space of $\mathbb{Z}$, denoted by $P_2(\mathbb{Z})$. Under the Assumption~\ref{ass:lp-bounded}, the target domain space is restricted to
\begin{equation}
\mathbb{D}(\mathbb{Z}) = \left\{D \in P_2(\mathbb{Z}) \;\middle|\; D \ll \nu, \; \frac{\mathrm{d}D}{\mathrm{d}\nu} \in L^2(\mathbb{Z}) \right\}.    
\end{equation}
We simply write it as $\mathbb{D}$. Let $\mathcal{P}$ be a meta-distribution over $\mathbb{D}$ satisfying
\begin{equation}
\nu = \mathbb{E}_{D \sim \mathcal{P}}[D].
\end{equation}
\end{Definition}


\begin{Definition}[\(C\)-regular domain generalization problem]
\label{def:C-regular}
Consider a fixed family of candidate target domains. Let \(\mathbb{Z}\) be a representation space satisfying Assumption~\ref{ass:lp-bounded} for this family, and let
\(\mathbb{D}_{\mathrm{tar}}(\mathbb{Z})\subseteq\mathbb{D}(\mathbb{Z})\) denote the measurable set of its induced domain distributions. Define
\begin{equation}
C(\mathbb{Z})
:=
\sup_{D\in\mathbb{D}_{\mathrm{tar}}(\mathbb{Z})}
\left\|
\frac{\mathrm{d}D}{\mathrm{d}\nu}
\right\|_{L^2(\nu)}^2.
\label{eq:C-def}
\end{equation}
We say that this family constitutes a \emph{\(C\)-regular domain generalization problem} with \(C:=C(\mathbb{Z})\) if
\(1<C<\infty\), \(\mathcal{P}(\mathbb{D}_{\mathrm{tar}}(\mathbb{Z}))>0\).
We then denote
\[
\mathbb{D}^{C}:=\mathbb{D}_{\mathrm{tar}}(\mathbb{Z}).
\]
The conditional meta-distribution of \(\mathcal{P}\) given \(\mathbb{D}^{C}\) is denoted by \(\mathcal{P}^{C}\) and is defined by
\begin{equation}
\mathcal{P}^{C}(A)
=
\frac{\mathcal{P}(A\cap\mathbb{D}^{C})}
     {\mathcal{P}(\mathbb{D}^{C})},
\qquad
A\subseteq\mathbb{D}\text{ measurable}.
\end{equation}
We denote the resulting problem by
\((\hat{\mathbb{D}}^{C}_{m},\mathbb{D}^{C})\),
where \(\hat{\mathbb{D}}^{C}_{m}\) denotes the sample of \(m\) source domains drawn independently from \(\mathcal{P}^{C}\).
\end{Definition}

We can use \(C\) to measure the cross-domain invariance captured by the representation space.

\begin{Definition}[Invariance measure captured by the representation space]
\label{def:cross-domain-invariance}
The cross-domain invariance captured by the representation space is defined as
\[
\mathrm{Inv}(\mathbb{Z}):=\frac{1}{\sqrt{C-1}}.
\]
\end{Definition}

\begin{Remark}
A larger \(\mathrm{Inv}(\mathbb{Z})\) means that the domain distributions are closer to the common reference \(\nu\), i.e., the representation space captures stronger cross-domain invariance.
\end{Remark}

Then, we define a functional class on domain space, and its Rademacher complexity term. 
\begin{Definition}[Risk Functional Class]
\label{def:dist-functional-space}
We define the risk functional induced by the loss function $\ell$ and a hypothesis $h \in \mathcal{H}$ as
\begin{equation}
\label{eq:phi-ell-h}
\phi_{\ell, h}(D) = E_D[\ell(h(z), f(z))],
\end{equation}
the class of risk functional induced by the hypothesis class $\mathcal{H}$ and loss function $\ell$ is defined as
\begin{equation}
\label{eq:Phi-ell-H}
\Phi_{\ell ,\mathcal{H}} = \{\phi_{\ell ,h}: \mathbb{D} \to \mathbb{R} \mid \phi_{\ell ,h}(D) = E_{D}[\ell (h(z), f(z))]; h\in \mathcal{H}\}.
\end{equation}
Its restriction on $\mathbb{D}^C$ is defined as
\begin{equation}
    \Phi_{\ell ,\mathcal{H}}^C = \{\phi_{\ell ,h}: \mathbb{D}^C \to \mathbb{R} \mid \phi_{\ell ,h}(D) = E_{D}[\ell (h(z), f(z))]; h\in \mathcal{H}\}.
\end{equation}
\end{Definition}

\begin{Definition}[Rademacher Complexity of Risk Functional Class]
\label{def:rademacher-dist}
The empirical Rademacher complexity of the risk functional class $\Phi_{\ell ,\mathcal{H}}$ with respect to the source domain distributions $D_1, \dots, D_m$ is defined as
\begin{equation}
\label{eq:rademacher-dist}
\hat{\mathcal{R}}_{D_1, \cdots, D_m}(\Phi_{\ell, \mathcal{H}})=E_{\sigma}[\sup_{\phi \in \Phi_{\ell, \mathcal{H}}}\frac{1}{m}\sum_{i=1}^m\sigma_i \phi(D_i)],
\end{equation}
where the components of $\boldsymbol{\sigma} = (\sigma_1,\dots,\sigma_m)$ are independent Rademacher variables (i.e., $\Pr(\sigma_i = +1) = \Pr(\sigma_i = -1) = 1/2$ for any $i \in \{1, \cdots, m\}$).
\end{Definition}

\subsection{McDiarmid-Type Measure Concentration Analysis on Wasserstein Space}
We leverage the distinctive geometric properties of the Wasserstein space to modify McDiarmid's inequality, thereby providing a Wasserstein-geometric confidence term for the distribution-level complexity, which directly characterizes the domain generalization performance in the error bound presented in the next subsection. We first introduce a new concept and assumption: 
\begin{Definition}[Multivariate Risk Functional]
\label{def:F}
For a $C$-regular domain generalization problem $(\hat{\mathbb{D}}^C_{m}, \mathbb{D}^C)$, we define a multivariate risk functional that takes its source domains as inputs:
\begin{equation}
\label{eq:F}
    F(D_1, \cdots, D_m)=\sup_{\phi \in \Phi_{\ell, \mathcal{H}}^C}E_{\mathcal{P}^C}[\phi(D)]-\frac{1}{m}\sum_{i=1}^m \phi(D_i),
\end{equation}
where $D_1,\dots,D_m \in \mathbb{D}^C$.
The class of all multivariate risk functionals is denoted as $\mathcal{F}^C$.
\end{Definition}


Then we modify the bounded differences condition in the McDiarmid's inequality to make it reflect the distinctive properties induced by the Wasserstein geometry.
\begin{Theorem}[Bounded Difference Condition for $\mathcal{F}^C$]
\label{thm:bounded-difference}
Under Assumptions~\ref{ass:lp-bounded}, \ref{CSA} and \ref{AssWD1}, and assuming the sample space $(\mathbb{Z},d,\nu)$ satisfies the $\mathrm{CD}(K(\mathbb{Z}),\infty)$ condition with $K(\mathbb{Z})>0$ (Definition~\ref{def:cdkinf} in Appendix~\ref{Add-Notation}), any multivariate risk functional $F \in \mathcal{F}^C$ satisfies the following bounded difference condition for any $k \in \{1,\dots,m\}$:
\begin{equation}
\label{eq:bounded-difference}
|F(D_1, \cdots, D_k, \cdots, D_m) - F(D_1, \cdots, D_k^{\prime}, \cdots, D_m)|\leq \frac{2\sqrt{2}}{m\cdot\mathrm{Inv}(\mathbb{Z})\cdot\sqrt{K(\mathbb{Z})}},
\end{equation}
where $K(\mathbb{Z})$ is the lower Ricci curvature bound of the representation space $\mathbb{Z}$.
\end{Theorem}

The proof of Theorem~\ref{thm:bounded-difference} can be found in Appendix~\ref{proof2}.

With the bounded difference property established, we can directly apply McDiarmid's inequality to obtain a concentration result for $F$ around its expectation.

\begin{Corollary}[Measure Concentration with Curvature Dependence]
\label{cor:concentration}
For any $t>0$, $F \in \mathcal{F}^C$ and event 
\begin{equation}\label{eq:event}
\begin{split}
    \Omega^m &=\big\{(D_1, \cdots, D_k, \cdots, D_m) \in (\mathbb{D}^C)^m:\\& \quad\quad F(D_1, \cdots, D_k, \cdots, D_m) - E_{(\mathcal{P}^c)^m}[F(D_1, \cdots, D_k, \cdots, D_m)]\geq t\big\},
\end{split}
\end{equation}
the following inequality holds:
\begin{equation}\label{eq:probexpbound}
    \int_{\Omega^m} {\rm d}(\mathcal{P}^C)^m  \leq 2\exp\left(-\frac{mt^2\cdot\mathrm{Inv}(\mathbb{Z})^2\cdot K(\mathbb{Z})}{4}\right),
\end{equation}
where $D_1, \cdots, D_m \in \mathbb{D}^C$.
\end{Corollary}
It shows that the deviation of the worst-case empirical risk (over the hypothesis class) from its expectation decays exponentially with the number of source domains $m$, at a rate that depends jointly on the lower curvature bound of the underlying representation space and the degree of invariance captured by it.

\subsection{A Domain Generalization Error Bound Incorporating Distribution‑Level Complexity and Wasserstein‑Geometric Confidence Term}\label{subsec_newB}
We now derive a high-probability upper bound on the domain generalization error $\operatorname{Err}_D(h)$, which incorporate a distribution‑level complexity and its Wasserstein‑geometric confidence term.
\begin{Theorem}[Domain Generalization Error Bound]
\label{thm:main-bound}
For a $C$-regular domain generalization problem $(\hat{\mathbb{D}}^C_{m}, \mathbb{D}^C)$, under Assumptions~
\ref{CSA} and \ref{AssWD1}, and assuming the sample space $(\mathbb{Z},d,\nu)$ satisfies the $\mathrm{CD}(K(\mathbb{Z}),\infty)$ condition with $K(\mathbb{Z})>0$, the following inequality holds uniformly for all $h \in \mathcal{H}$ with probability at least $1 - 3\delta$ over the random sampling of the source domains $D_1,\dots,D_m \sim \mathcal{P}^C$ and the intra-domain samples $S_i = \{x_{i,j}\}_{j=1}^n$:
    \begin{equation}\label{eq:main-bound}
    \small
        \sup_{h \in \mathcal{H}}Err_D(h) \leq 2\underbrace{\hat{\mathcal{R}}_{D_1, \cdots, D_m}(\Phi_{\ell, \mathcal{H}})}_{\text{Complexity term 1}} + \frac{2q}{m}\underbrace{\sum_{i=1}^m \mathcal{\hat{R}}^i_n(\mathcal{H})}_{\text{Complexity term 2}} + \underbrace{\mathcal{O}\Big(\sqrt{\frac{\ln(2/\delta)}{n}}\Big)}_{\text{Confidence term 1}}+\underbrace{\mathcal{O}\Big( \sqrt{\frac{\ln(2/\delta)}{m\cdot\mathrm{Inv}(\mathbb{Z})^2\cdot K(\mathbb{Z})}}\Big)}_{\text{Confidence term 2}}. 
    \end{equation}
Here $\mathcal{\hat{R}}^i_n(\mathcal{H})$ is the Rademacher complexity term with respect to samples of $i$-th source domain.
\end{Theorem}

The proof of Theorem~\ref{thm:main-bound} can be found in Appendix~\ref{proof3}.

The derived bound consists of four terms: Complexity term 1 is the Rademacher complexity of the risk functional class with respect to the source domain distributions, which shows that distribution shift introduces an additional complexity penalty; Complexity term 2 is the Rademacher complexity computed on the samples from all source domains and reflects the degree to which the model fits the training samples; Confidence term 1 is a standard confidence term in generalization analysis that decays as $1/\sqrt{n}$, i.e., inversely proportional to the square root of the total number of samples; and Confidence term 2 is an additional confidence term arising from the distribution-level randomness. It depends on the lower Ricci curvature bound $K(\mathbb{Z})$ and the invariance degree $\mathrm{Inv}(\mathbb{Z})$ of the representation space; a larger $K(\mathbb{Z})$ reduces the term, implying that learned representation spaces with a larger lower bound on Ricci curvature are more favorable for domain generalization.


\section{Experiment}\label{Expe}
In this section, we conduct two sets of experiments to validate our conclusions:
(1) whether latent space curvature affects domain generalization, and (2) whether
the number of source domains influences generalization performance. We adopt an
ERM encoder-classifier framework and evaluate on PACS, Office-Home, and
Digits-DG, with dataset details provided in Appendix~\ref{exp2}. Following the
leave-one-domain-out protocol in \cite{Lid31}, \cite{Carlucci34}, and
\cite{lida37}, one domain is used for testing and the remaining domains for
training. We set the batch size to 64 and average results over three random
seeds using the Dassl.pytorch toolbox. Due to space limitations, we present the
PACS results here; the Office-Home and Digits-DG results are provided in
Appendix~\ref{expD}.

\begin{figure*}[t]
	\centering
    \subfigure{
		\begin{minipage}[t]{0.23\linewidth}
			\centering			\includegraphics[width=1\linewidth]{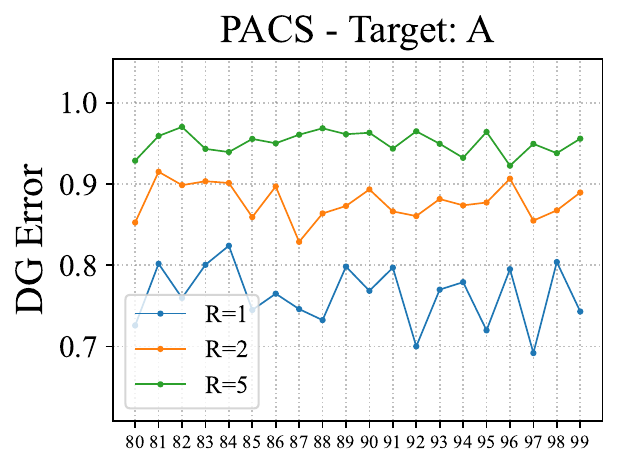}           
		\end{minipage}
	}
    \subfigure{
		\begin{minipage}[t]{0.23\linewidth}
			\centering
			\includegraphics[width=1\linewidth]{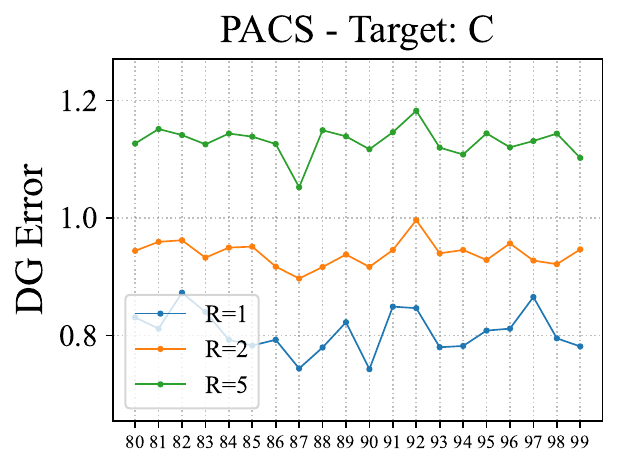}            
		\end{minipage}
	}
    \subfigure{
		\begin{minipage}[t]{0.23\linewidth}
			\centering
			\includegraphics[width=1\linewidth]{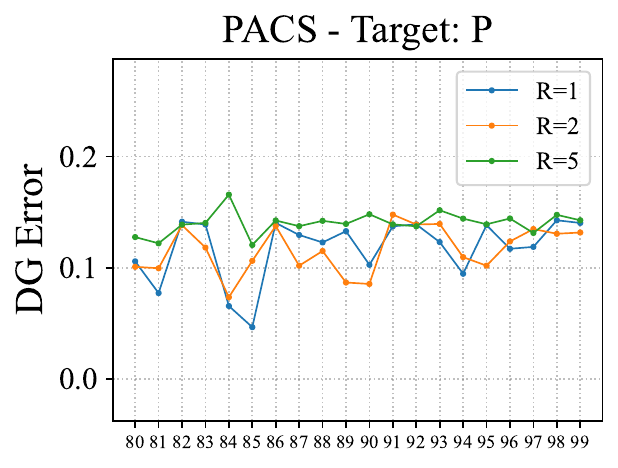}            
		\end{minipage}
	}
    \subfigure{
		\begin{minipage}[t]{0.23\linewidth}
			\centering
			\includegraphics[width=1\linewidth]{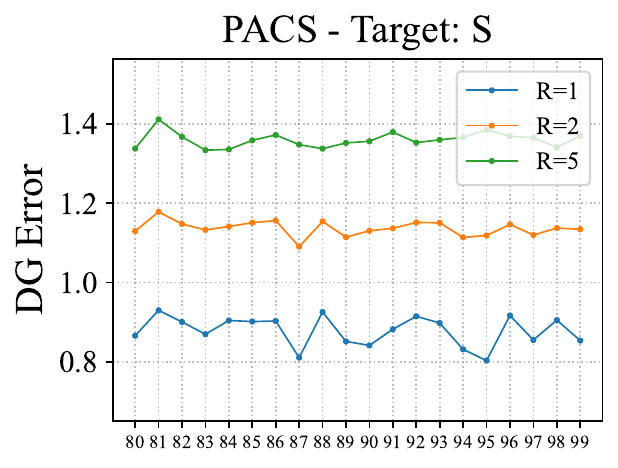}            
		\end{minipage}
	}
    \subfigure{
		\begin{minipage}[t]{0.23\linewidth}
			\centering			\includegraphics[width=1\linewidth]{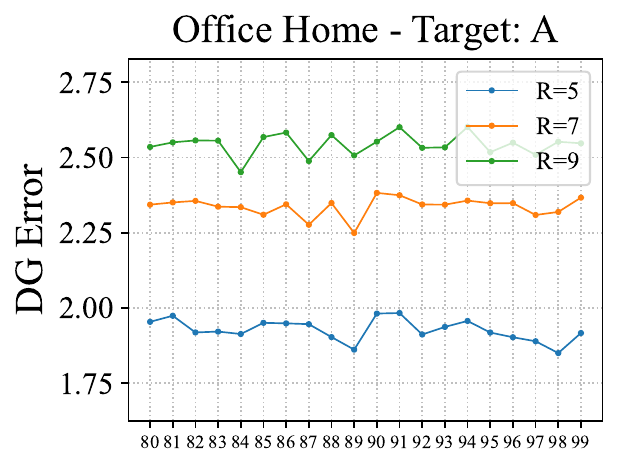}           
		\end{minipage}
	}
    \subfigure{
		\begin{minipage}[t]{0.23\linewidth}
			\centering
			\includegraphics[width=1\linewidth]{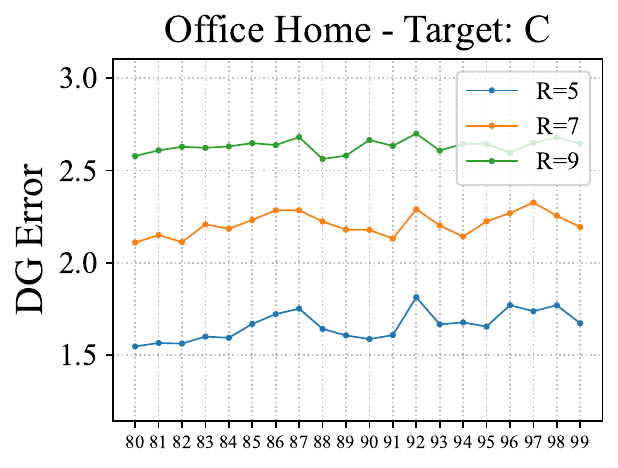}            
		\end{minipage}
	}
    \subfigure{
		\begin{minipage}[t]{0.23\linewidth}
			\centering
			\includegraphics[width=1\linewidth]{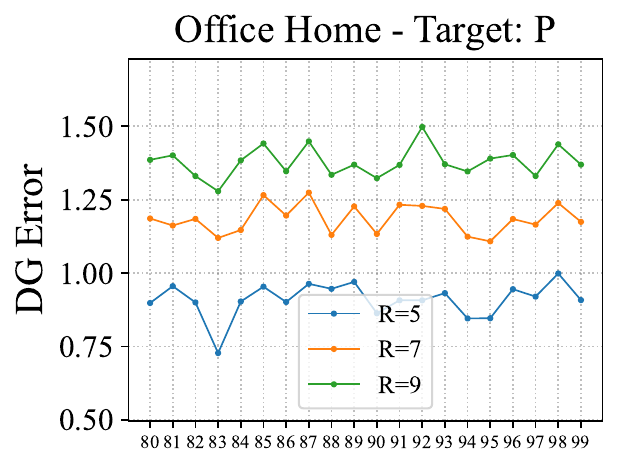}            
		\end{minipage}
	}
    \subfigure{
		\begin{minipage}[t]{0.23\linewidth}
			\centering
			\includegraphics[width=1\linewidth]{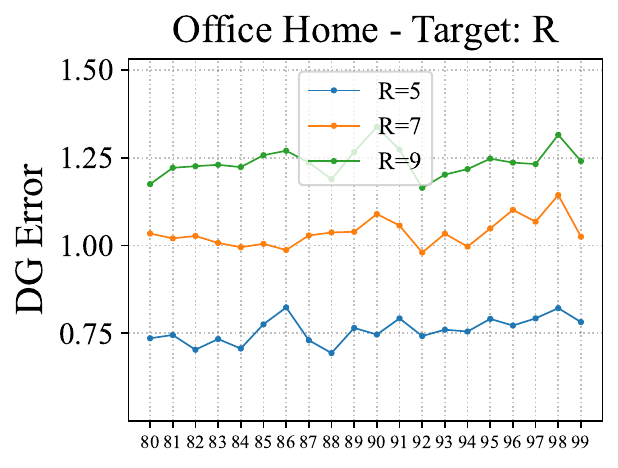}            
		\end{minipage}
	}
    \subfigure{
		\begin{minipage}[t]{0.23\linewidth}
			\centering			\includegraphics[width=1\linewidth]{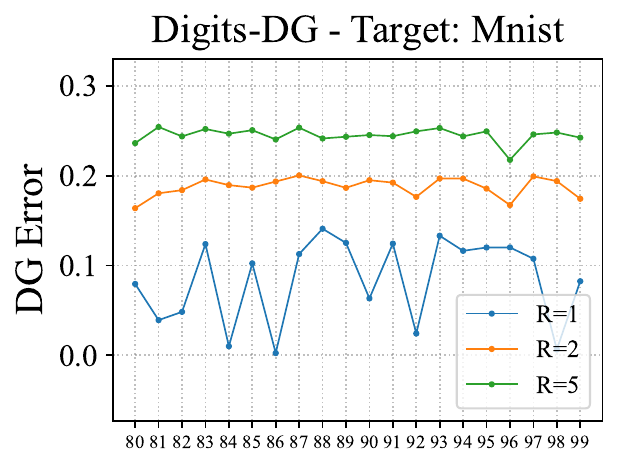}           
		\end{minipage}
	}
    \subfigure{
		\begin{minipage}[t]{0.23\linewidth}
			\centering
			\includegraphics[width=1\linewidth]{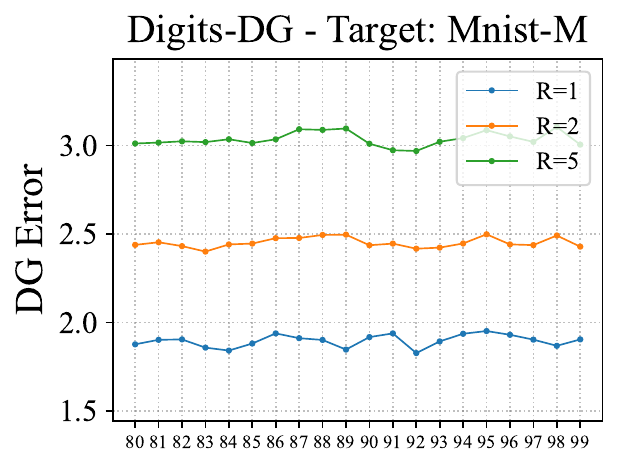}            
		\end{minipage}
	}
    \subfigure{
		\begin{minipage}[t]{0.23\linewidth}
			\centering
			\includegraphics[width=1\linewidth]{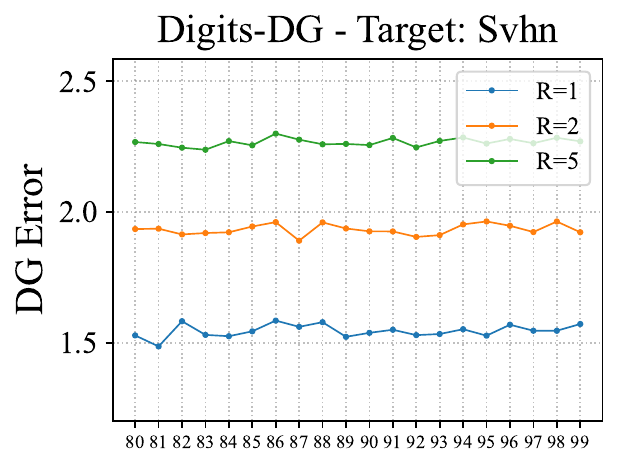}            
		\end{minipage}
	}
    \subfigure{
		\begin{minipage}[t]{0.23\linewidth}
			\centering
			\includegraphics[width=1\linewidth]{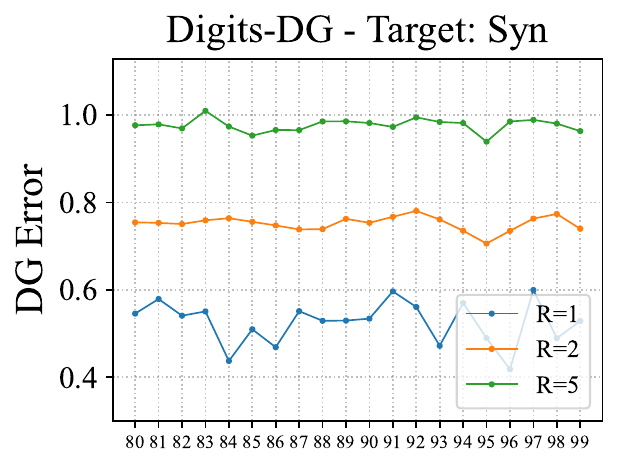}            
		\end{minipage}
	}
	\caption{Comparison of domain generalization errors in latent spaces with different curvatures on the PACS, Office-Home and Digits-DG datasets}
	\label{fig1}
\end{figure*}

\subsection{Impact of Latent Space Curvature for Domain Generalization}\label{exp4.1}
Our aim is to study the impact of latent space curvature on domain generalization.
To this end, we require a manifold with controllable curvature and a probability
distribution defined on its geometric support. A natural choice is the
von Mises--Fisher (vMF) distribution, which is defined on a hyperspherical
geometry and uses a single radius parameter $R$ to control curvature. We further
extend this framework to a mixture of multiple hyperspheres, each with
equal curvature but learnable centers, which substantially increases
representational capacity while retaining efficient curvature control.

In our framework, an encoder maps each input image $x$ to the parameters
$(\boldsymbol{\mu}, \kappa)$ of a vMF distribution, and a latent code
$\mathbf{z}_{\text{unit}} \sim \mathrm{vMF}(\boldsymbol{\mu}, \kappa)$ is sampled
on the unit hypersphere. To avoid the limited capacity of a single small-radius
sphere, we construct a latent space as a mixture of $N_{\text{sph}}$
hyperspheres sharing radius $R$ but having distinct learnable centers
$\{\mathbf{c}_i\}_{i=1}^{N_{\text{sph}}}$. A Gumbel--Softmax assignment
$\boldsymbol{\pi}$ selects the contribution of each sphere, and the final
representation is
\begin{equation}
\mathbf{z} = R \cdot \mathbf{z}_{\text{unit}}
+ \sum_{i=1}^{N_{\text{sph}}} \pi_i \mathbf{c}_i .
\end{equation}
The curvature of each individual sphere remains $\mathcal{C} = 1/R^2$, so
controlling $R$ directly governs the geometric regularization of the
representation. The scaled latent code $\mathbf{z}$ is then passed to a
classifier for label prediction, and the entire pipeline is trained end-to-end
with a supervised classification loss. Detailed descriptions of the vMF
distribution, network architecture, hyperparameter settings, and training
procedure are provided in Appendix~\ref{app:impl_details 1}.

Figure~\ref{fig1} present the
experimental results on three benchmark datasets: PACS, Office-Home, and
Digits-DG. Our results largely corroborate our theoretical findings: in most
cases, increasing the latent space curvature (i.e., decreasing the
hyperspherical radius) leads to a smaller domain generalization error. This trend
is consistently observed across various target domains on PACS, with one notable
exception: when Photo is used as the target domain. In this setting, the domain
generalization error remains consistently low across all curvature values, and no
clear relationship between curvature and generalization performance is evident.
A detailed discussion of this anomaly is deferred to
Appendix~\ref{app:impl_details 1}.

\subsection{Impact of Source Domain Number on Domain Generalization}
\label{exp4.2}

We further investigate the relationship between the number of source domains and
the domain generalization error. To this end, we design a controlled experiment
that systematically varies the number and composition of source domains used
during training, while keeping all other experimental settings fixed. For PACS,
Office-Home, and Digits-DG, we consider four source configurations: a single
source domain, two source domains, three source domains (the standard
multi-source setting), and six source domains obtained by applying MixStyle
\citep{Zhou52} to mix feature statistics of source-domain pairs. For each
dataset and each target domain, we train a pretrained ResNet-18 backbone with a
classification head, and measure the domain generalization error as the test
error on the target domain minus the average training error on the source
domains.

\begin{figure*}[t]
	\centering
    \subfigure{
		\begin{minipage}[t]{0.23\linewidth}
			\centering			\includegraphics[width=1\linewidth]{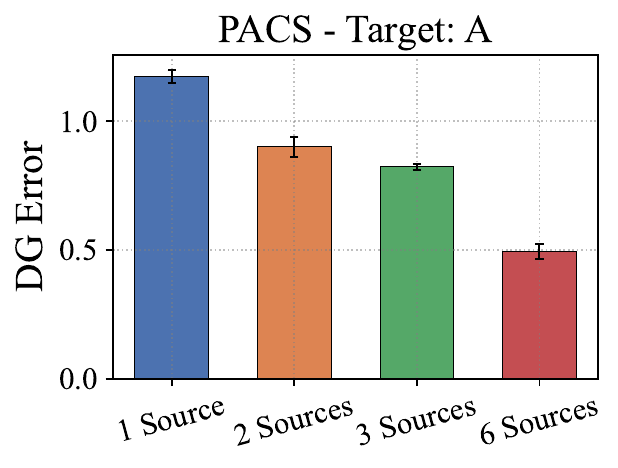}           
		\end{minipage}
	}
    \subfigure{
		\begin{minipage}[t]{0.23\linewidth}
			\centering
			\includegraphics[width=1\linewidth]{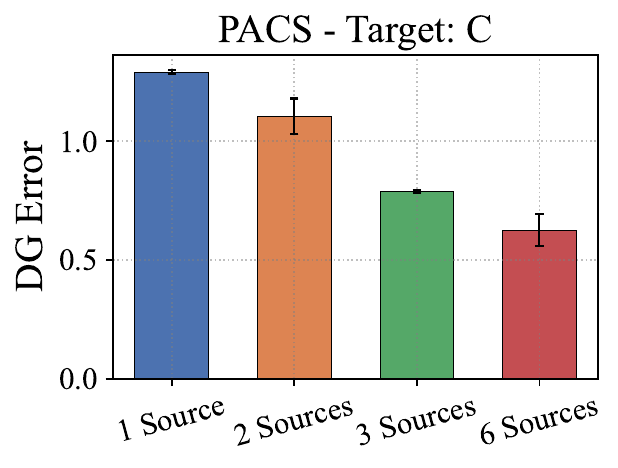}            
		\end{minipage}
	}
    \subfigure{
		\begin{minipage}[t]{0.23\linewidth}
			\centering
			\includegraphics[width=1\linewidth]{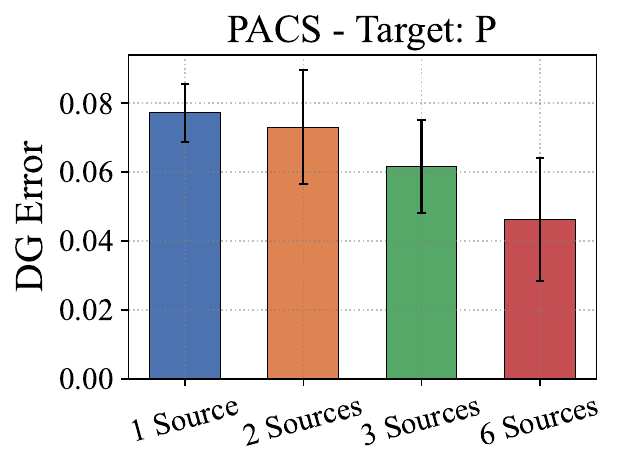}            
		\end{minipage}
	}
    \subfigure{
		\begin{minipage}[t]{0.23\linewidth}
			\centering
			\includegraphics[width=1\linewidth]{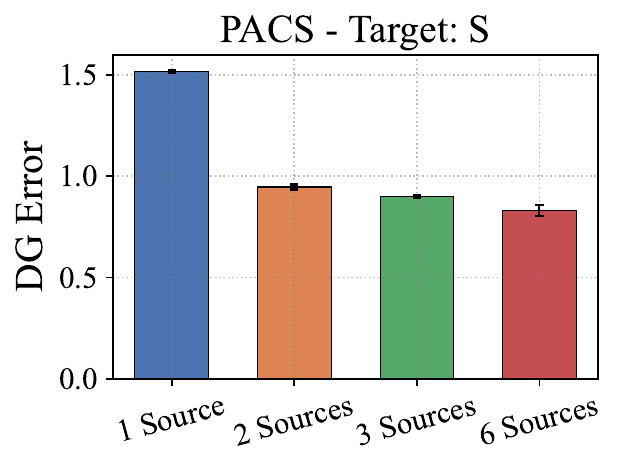}            
		\end{minipage}
	}
    \subfigure{
		\begin{minipage}[t]{0.23\linewidth}
			\centering			\includegraphics[width=1\linewidth]{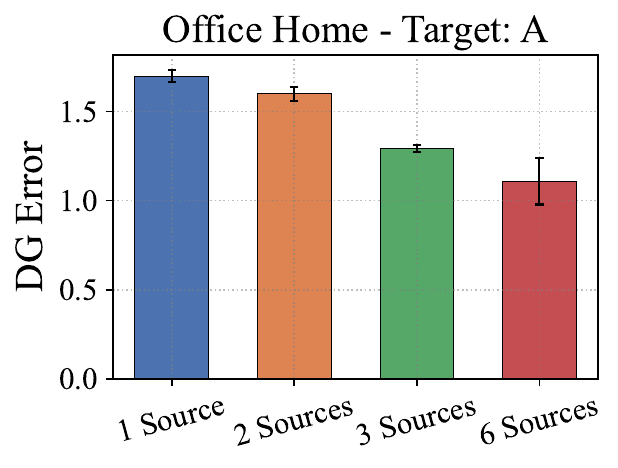}           
		\end{minipage}
	}
    \subfigure{
		\begin{minipage}[t]{0.23\linewidth}
			\centering
			\includegraphics[width=1\linewidth]{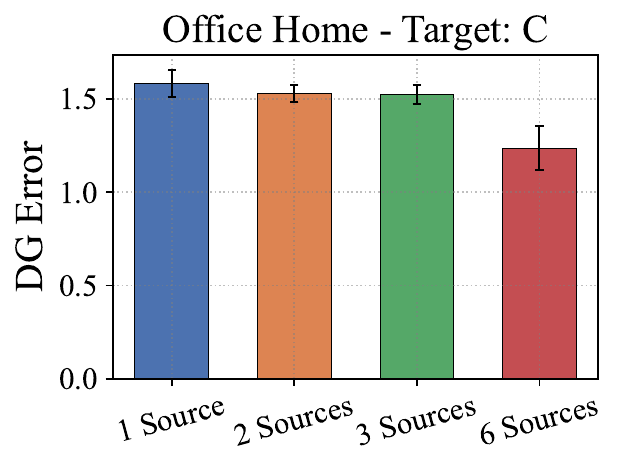}            
		\end{minipage}
	}
    \subfigure{
		\begin{minipage}[t]{0.23\linewidth}
			\centering
			\includegraphics[width=1\linewidth]{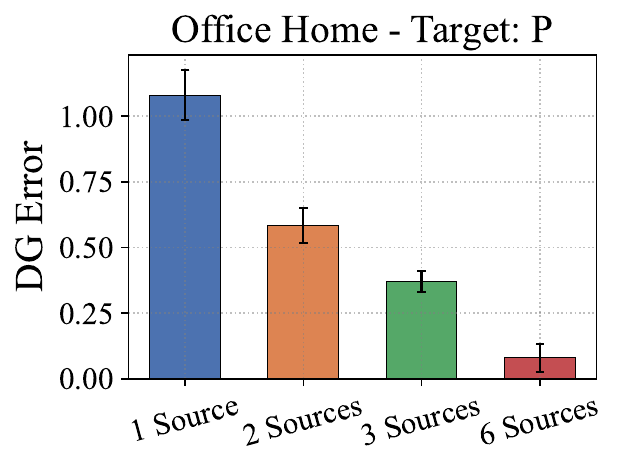}            
		\end{minipage}
	}
    \subfigure{
		\begin{minipage}[t]{0.23\linewidth}
			\centering
			\includegraphics[width=1\linewidth]{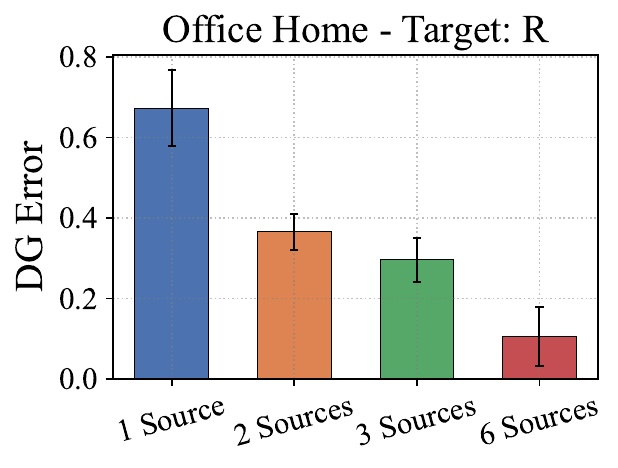}            
		\end{minipage}
	}
        \subfigure{
		\begin{minipage}[t]{0.23\linewidth}
			\centering			\includegraphics[width=1\linewidth]{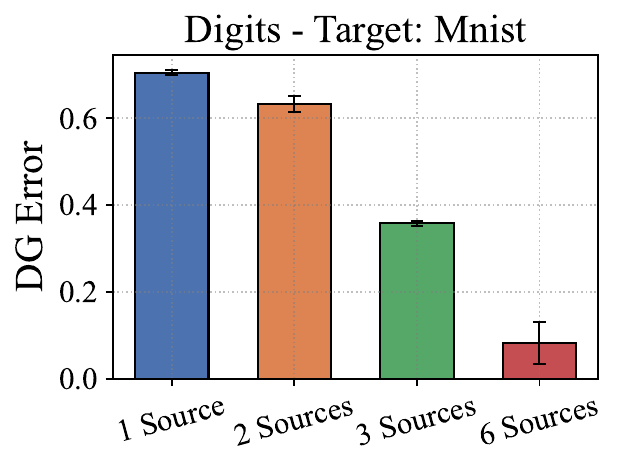}           
		\end{minipage}
	}
    \subfigure{
		\begin{minipage}[t]{0.23\linewidth}
			\centering
			\includegraphics[width=1\linewidth]{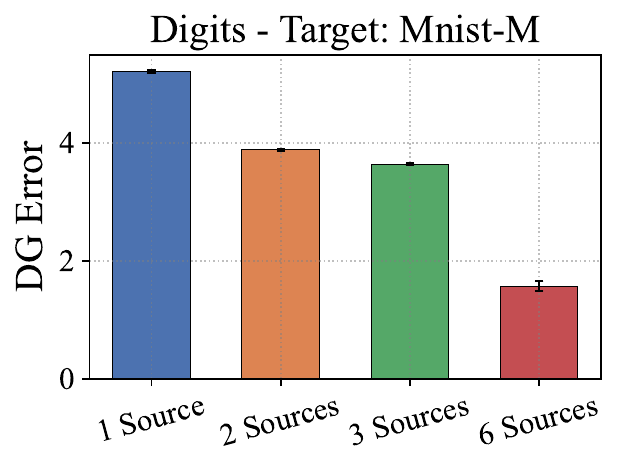}            
		\end{minipage}
	}
    \subfigure{
		\begin{minipage}[t]{0.23\linewidth}
			\centering
			\includegraphics[width=1\linewidth]{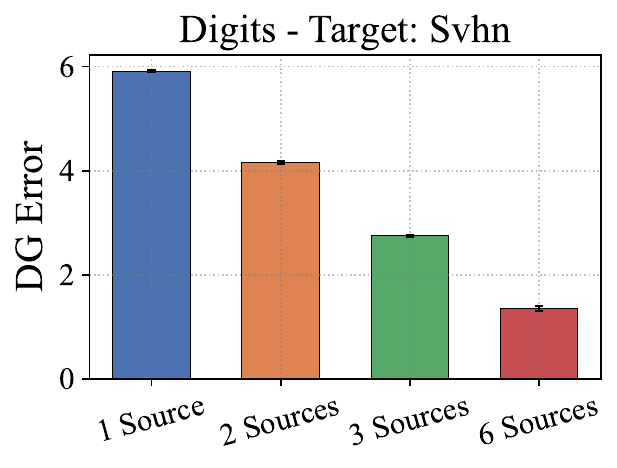}            
		\end{minipage}
	}
    \subfigure{
		\begin{minipage}[t]{0.23\linewidth}
			\centering
			\includegraphics[width=1\linewidth]{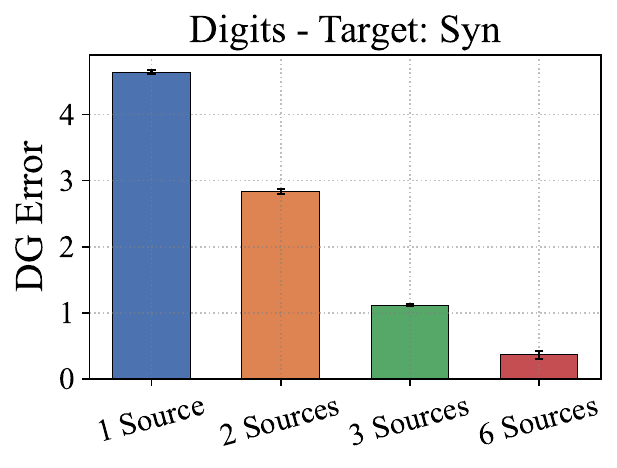}            
		\end{minipage}
	}
	\caption{Comparison of domain generalization errors with different source domain number on the PACS, Office-Home and Digits-DG datasets}
	\label{fig12}
\end{figure*}

Figure~\ref{fig12} reports the results as bar charts. In most cases, the results are consistent with our
theoretical bound: increasing the number of source domains generally reduces the
domain generalization error. However, two exceptions are observed: when Photo is
used as the target domain on PACS, the results exhibit notable random
fluctuations, and when Clipart is used as the target domain on Office-Home, the
influence of the number of source domains on the generalization error is less
pronounced. We provide possible explanations for these exceptions in Appendix~\ref{app:impl_details 2}.

\section{Conclusion}\label{Conc}
This paper addresses the fundamental limitation of existing geometry-free theories of domain generalization, which treat the domain space as devoid of geometry and thus fail to formally characterize distribution-level generalization under distribution shifts. We formalize the domain space as the Wasserstein space over the representation space and, building on its \(\mathrm{CD}(K,\infty)\) curvature-dimension geometry, derive a novel geometry-dependent domain generalization error bound. By introducing a curvature-dependent confidence term, this bound formally disentangles sample-level and distribution-level generalization effects and reveals that domain generalization is governed by both the invariance and the curvature of the representation space. Finally, experimental results validate the effect of curvature on domain generalization, supporting the theoretical findings.

\subsection*{Ai Use Statement}
In this work, we used generative AI tools to improve the readability and clarity of the manuscript and to assist with software code development. All AI-assisted edits and code were carefully reviewed, verified, and revised by the authors when necessary. We take responsibility for the final content of this work.

\enlargethispage{2\baselineskip}


\bibliography{iclr2027_conference2}
\bibliographystyle{iclr2027_conference}

\appendix
\section{Related Work}\label{RWork}
In this appendix, we introduce related work to provide a more comprehensive understanding of this research direction. The following topics are covered in order: classical generalization theory and domain generalization.
\subsection{Generalization}
To quantify the generalization of machine learning models, numerous foundational theoretical frameworks have been established over the past decades. 
In addition to the classical works in this field introduced in Section~\ref{intro}, we further review several studies that analyze generalization from different perspectives. Among them, the use of information-theoretic tools to analyze generalization error constitutes an important direction.
\cite{xu2017information} proposes an information-theoretic framework based on mutual information to quantify and control the bias induced by the selection process in adaptive data analysis, and demonstrates that this mutual-information bound provides tight and actionable upper bounds on bias in a variety of common analytical settings.
\cite{bu2019tightening, harutyunyan2021information, haghifam2021towards} extended it by using different conditional versions of mutual information.
Moreover, some works such as \cite{arora2018stronger, barsbey2021heavy} introduce a compression‑based generalization framework. These studies show that a higher level of compressibility indicates a lower generalization error, because the hypothesis class $\mathcal{W}$ can be approximated by a smaller “compressed” space, which intuitively leads to a lower worst-case error.
Recently, \cite{simsekli2020hausdorff} proves that the Hausdorff dimension of SGD trajectories is governed by the heavy-tailedness of the gradient noise, and derives a generalization error bound based on this fractal dimension, thereby revealing that heavy-tailed noise enhances generalization by reducing the effective geometric complexity of the parameter trajectories as an implicit regularization mechanism.
Finally, by unifying mutual information, compressibility, and fractal dimensions within a single rate-distortion framework, \cite{sefidgaran2022rate} innovatively treats generalization error as a form of lossy distortion and requires compressibility to hold within any KL-divergence neighborhood of the true data distribution, thereby deriving tighter expected and high-probability generalization bounds.

\subsection{Domain Generalization} 
The first formal characterization of the DG problem setting was provided by \cite{Blanchard79}. 
It worked within a reproducing kernel Hilbert space for the hypothesis class, learned from multiple related classification tasks, and established a distribution-free generalization error analysis. The subsequent work of \cite{Blanchard08} further introduced a marginal transfer learning framework, which incorporates marginal distributions into the feature space to explicitly address domain shift.
\cite{Li37} posited a linear dependence in the latent feature space across domains, suggesting that any domain can be expressed as a linear combination of the others.
\cite{Ye44} introduced a two-component quantification of feature distribution: a variation term capturing cross-domain stability and an informativeness term reflecting discriminative power. Using an expansion function driven by the variation term, they bounded the error on unseen domains, with the informativeness term indicating learnability. Although this approach relaxes prior assumptions, its resulting domain generalization error bound suffers from limited interpretability and offers little practical guidance.
\cite{wang2026bridging} highlights the complementarity of the two different domain generalization strategies by decomposing the feature space into different subspaces and analyzing domain generalization methods based on domain-invariant representation learning and those based on domain augmentation within these subspaces.

\section{Covariate Shift Setting}
\label{App_CSA}
The covariate shift is the most general setting in domain generalization and is widely applied in domain generalization works. It assumes that all domains adhere to identical classification rules, thereby ensuring consistency in their labeling functions. Based on this assumption, any distribution shift can be attributed to potential differences in the marginal feature distributions across individual domains, which is referred to as covariate shift. We use the following formula to represent:\\
\begin{Assumption}[Covariate Shift Setting]\label{CSA}
    For any two domains $D, D^{\prime} \in \mathbb{D}$, let $f_D$ and $f_{D^{\prime}}$ be their respective labeling functions. Then these functions satisfy:
\begin{equation}
f_D(z) = f_{D^{\prime}}(z), \quad \forall z \in \mathbb{Z}.
\end{equation}
\end{Assumption}

\section{Wasserstein Geometry}\label{Add-Notation}
In this appendix, we introduce the geometric tools from optimal transport that are needed to analyze domain distributions as fundamental objects. 
Unlike classical learning theory, which operates on a fixed sample space, our analysis treats each domain as a probability distribution and the collection of all domains as a subset of the $2$-Wasserstein space. 
This space carries a rich metric structure, but its probabilistic behavior—such as the concentration of distribution-valued random variables—is governed by curvature. 
\begin{Definition}[Metric-measure Space and Wasserstein Space]
\label{def:msws}
Let $\mathbb{X}$ be a compact length space and let $\nu$ be a Borel probability measure on $\mathbb{X}$, referred to as the reference measure. The triple $(\mathbb{X}, d, \nu)$ is called a metric-measure space, where $d$ is the distance of $\mathbb{X}$.

Let $P(\mathbb{X})$ be the set of all Borel probability measures on $\mathbb{X}$,
For any two measures $D_1, D_2 \in P(\mathbb{X})$, the $p$-Wasserstein distance is defined as
\begin{equation}\label{wasasitan}
    W_p(D_1,D_2) = \inf_{\gamma \in \Pi(D_1,D_2)} \Big(\int_{\mathbb{X} \times \mathbb{X}} c(x_1, x_2)^p d\gamma(x_1,x_2)\Big)^{\frac{1}{p}},
\end{equation}
where $\Pi(D_1,D_2)$ is a collection of all joint probability measures on $\mathbb{X}\times\mathbb{X}$ with marginals $D_1$ and $D_2$, $c(x_1, x_2)$ is a cost function for transporting one unit of mass $x_1$ to $x_2$.
Equipped with the $2$-Wasserstein distance $W_2(\cdot, \cdot)$, $P(\mathbb{X})$ becomes a metric space, denoted by $P_2(\mathbb{X})$ and called the $2$-Wasserstein space.
\end{Definition}

\begin{Definition}[CD(K,$\infty$) curvature condition \citep{Lott91}]
\label{def:cdkinf}
A metric measure space $(\mathbb{X}, d, \nu)$ is said to have Ricci curvature bounded below by $K \in \mathbb{R}$ (or to satisfy the $\mathrm{CD}(K,\infty)$ condition) if for every pair $\mu_0, \mu_1 \in P_2(\mathbb{X}, \nu)$, where $P_2(\mathbb{X}, \nu)=\{\mu \in P_2(\mathbb{X}) : \operatorname{supp}(\mu) \subset \operatorname{supp}(\nu)\}$, there exists a Wasserstein geodesic $\{\mu_t\}_{t \in [0,1]}$ connecting them such that the relative entropy $H_{\infty, \nu}(\mu) = \int_\mathbb{X} \rho \log \rho {\rm d}\nu$ (where $\rho = {\rm d}\mu/ {\rm d}\nu$) satisfies the displacement $K$-convexity inequality along this geodesic:
\begin{equation}
\label{eq:displacement-k-convexity}
H_{\infty, \nu}(\mu_{t}) \leq (1 - t)H_{\infty, \nu}(\mu_{0}) + tH_{\infty, \nu}(\mu_{1}) - \frac{K}{2} t(1 - t)W_{2}(\mu_{0},\mu_{1})^{2}, \quad \forall t\in [0,1].
\end{equation}
\end{Definition}

\section{Brief Introduction to \cite{Redko93}}
\begin{Assumption}\label{AssWD1}
The cost function takes the form $c(x_1, x_2) = \| \phi(x_1) - \phi(x_2) \|_{\mathcal{H}_{k_l}}$, where $\mathcal{H}_{k_l}$ is a reproducing kernel Hilbert space (RKHS) with kernel $k_l$ induced by the feature map $\phi$. Moreover, $k_l$ is assumed to be square-root integrable with respect to $D_1$ and $D_2$ in Eqn.~\ref{wasasitan}, and satisfies $0 \leq k_l(x_1,x_2) \leq L$. The loss function $\ell(h(x), f(x))$ is convex, symmetric, bounded, satisfies the triangle inequality, and admits the parametric form $|h(x) - f(x)|^p$ for some $p>0$.
\end{Assumption}
Under these assumptions, the following domain adaptation bound based on the first-order Wasserstein distance is established.
\begin{Theorem}[Domain Adaptation Risk Bound \cite{Redko93}]
\label{thm:wasserstein_error}
Under Assumptions~\ref{AssWD1}, for any two hypotheses $h, h'$, the difference in their expected risks between the target and source domains is bounded by the first-order Wasserstein distance between the domain distributions:
\begin{equation}
|\mathcal{E}_{D_1}(h,h^{\prime}) - \mathcal{E}_{D_2}(h,h^{\prime})| \leq W_{1}(D_1, D_2),
\end{equation}
where $\mathcal{E}_{D}(h, h')$ denotes the expected risk of the hypothesis $h$ with respect to $h'$ under distribution $D$, and $W_1(\cdot, \cdot)$ is the Wasserstein distance of order $1$.
\end{Theorem}

\section{Proof of Theorem~\ref{thm:bounded-difference}}\label{proof2}
\begin{proof}
For a $C$-regular domain generalization problem $(\hat{\mathbb{D}}^C_{m}, \mathbb{D}^C)$, we have $D_1, \dots D_m \in \mathbb{D}^C$. 
We consider the change in the functional \(F \in \mathcal{F}^C\) when the \(k\)-th distribution \(D_k\) is replaced by \(D_k^\prime \in \mathbb{D}^C\). By definition,
\begin{equation}
\begin{split}
    &\quad F(D_1, \cdots, D_k, \cdots, D_m) - F(D_1, \cdots, D_k^{\prime}, \cdots, D_m)\\&=
    \sup_{\phi \in \Phi^C_{\ell, \mathcal{H}}} E_{\mathcal{P}^C}[\phi(D)]-\frac{1}{m}\sum_{i=1}^m\phi(D_i)\\&\quad-\sup_{\phi \in \Phi^C_{\ell, \mathcal{H}}}E_{\mathcal{P}^C}[\phi(D)]-\frac{1}{m}\big(\sum_{i\neq k}\phi(D_i)+\phi(D_k^{\prime})\big)
    \\&\leq \sup_{\phi \in \Phi^C_{\ell, \mathcal{H}}}\frac{\phi(D_k)-\phi(D_k^{\prime})}{m}\\&=\frac{1}{m}\sup_{h \in \mathcal{H}}E_{D_k}[\ell(h(z), f(z))]-E_{D_k^{\prime}}[\ell(h(z), f(z))]\\& \leq \frac{1}{m}\sup_{h \in \mathcal{H}}\big|E_{D_k}[\ell(h(z), f(z))]-E_{D_k^{\prime}}[\ell(h(z), f(z))]\big|.
\end{split}
\end{equation}

Let $h^{\ast}=\operatorname{argsup}_{h \in \mathcal{H}} \big|E_{D_k}[\ell(h(z), f(z))]-E_{D_k^{\prime}}[\ell(h(x), f(z))]\big|$, then 
\begin{equation}
    \begin{split}
        &\quad F(D_1, \cdots, D_k, \cdots, D_m) - F(D_1, \cdots, D_k^{\prime}, \cdots, D_m)\\& \leq \frac{1}{m}\sup_{h \in \mathcal{H}}\big|E_{D_k}[\ell(h(z), f(z))]-E_{D_k^{\prime}}[\ell(h(z), f(z))]\big|\\&= \frac{1}{m}
        \big|\mathcal{E}_{D_k}(h^{\ast},f) - \mathcal{E}_{{D_k}^{\prime}}(h^{\ast},f)\big|\\& \leq \frac{W_1(D_k, D_k^{\prime})}{m}, 
    \end{split}
\end{equation}
the inequality in the fourth line is due to Theorem~\ref{thm:wasserstein_error}.

Due to the Hölder inequality, for any $\gamma \in \Pi(D_k, D_k^{\prime})$ and cost function $c(\cdot, \cdot)$, we have 
\begin{equation}
    \begin{split}
        \int_{\mathbb{Z} \times \mathbb{Z}} c(z_1,z_2) d\gamma(z_1,z_2) &\leq \Big(\int_{\mathbb{Z} \times \mathbb{Z}} c(z_1,z_2)^2 d\gamma(z_1,z_2)\Big)^{\frac{1}{2}} \cdot \Big(\int_{\mathbb{Z} \times \mathbb{Z}} 1^2d\gamma(z_1,z_2)\Big)^{\frac{1}{2}}\\&=\Big(\int_{\mathbb{Z} \times \mathbb{Z}} c(z_1,z_2)^2 d\gamma(z_1,z_2)\Big)^{\frac{1}{2}}
    \end{split}
\end{equation}
Taking the infimum on both sides yields:
\begin{equation}
    \begin{split}
        W_1(D_k, D_k^{\prime})&=\inf_{\gamma \in \Pi(D_k, D_k^{\prime})}\int_{\mathbb{Z} \times \mathbb{Z}} c(z_1,z_2) d\gamma(z_1,z_2)\\&\leq \inf_{\gamma \in \Pi(D_k, D_k^{\prime})} \Big(\int_{\mathbb{Z} \times \mathbb{Z}} c(z_1,z_2)^2 d\gamma(z_1,z_2)\Big)^{\frac{1}{2}}\\&= \Big(\inf_{\gamma \in \Pi(D_k, D_k^{\prime})}\int_{\mathbb{Z} \times \mathbb{Z}} c(z_1,z_2)^2 d\gamma(z_1,z_2)\Big)^{\frac{1}{2}}\\&=W_2(D_k, D_k^{\prime})
    \end{split}
\end{equation}
So we have 
\begin{equation}\label{eqFD1}
    \begin{split}
        &\quad F(D_1, \cdots, D_k, \cdots, D_m) - F(D_1, \cdots, D_k^{\prime}, \cdots, D_m)\\& \leq \frac{W_1(D_k, D_k^{\prime})}{m} \leq \frac{W_2(D_k, D_k^{\prime})}{m} \\&\leq \frac{W_2(D_k, \nu)+W_2(D_k^{\prime}, \nu)}{m}\\&\leq \frac{1}{m}\Big(\sqrt{\frac{2H_{\infty, \nu}(D_k)}{K(\mathbb{Z})}}+\sqrt{\frac{2H_{\infty, \nu}(D_k^\prime)}{K(\mathbb{Z})}}\Big),
    \end{split}
\end{equation}
The inequality in the last line is due to the Theorem~6.1 in \cite{Lott91}.

Similarly we can obtain:
\begin{equation}\label{eqFD2}
\begin{split}
    &\quad F(D_1, \cdots, D_k^{\prime}, \cdots, D_m) - F(D_1, \cdots, D_k, \cdots, D_m)\\&\leq \frac{1}{m}\Big(\sqrt{\frac{2H_{\infty, \nu}(D_k)}{K(\mathbb{Z})}}+\sqrt{\frac{2H_{\infty, \nu}(D_k^\prime)}{K(\mathbb{Z})}}\Big).
    \end{split}
\end{equation}
Let $\rm{d}D = \rho \rm{d}\nu$, we have:
\begin{equation}\label{eqFD3}
    \begin{split}
        H_{\infty, \nu}(D)&=\int_\mathbb{X}\rho\log\rho{\rm d}\nu\\&= \int_\mathbb{X}\log\rho{\rm d}D\\&\leq \log\Big(\int_\mathbb{X}\rho{\rm d}D\Big)\\& =\log\Big(\int_\mathbb{X}\rho^2{\rm d}\nu\Big)\\&\leq \int_\mathbb{X}\rho^2{\rm d}\nu-1\\&=\|\rho\|^2_{L^2}-1.
    \end{split}
\end{equation}
The first ``$\leq$'' is due to the $\log(\cdot)$ is a concave function and the Jensen’s inequality. The second ``$\leq$'' is due to the $\log{x}\leq x-1$.

Because the $D \in \mathbb{D}^C$, $\|\rho\|^2_{L^2}\leq C$, so we have:
\begin{equation}
    H_{\infty, \nu}(D)\leq C-1.
\end{equation}
According to Hölder’s inequality, we have: 
\begin{equation}
1 = \int \rho \mathrm{d}\nu \le \left( \int \rho^2 \mathrm{d}\nu \right)^{1/2} = \|\rho\|_{L^2} \leq \sqrt{C},
\end{equation}
so 
\begin{equation}
    C-1 \geq 0.
\end{equation}
Letting $\mathrm{Inv}(\mathbb{Z}):=\frac{1}{\sqrt{C-1}}$, and combining Eqn.~\ref{eqFD1}, Eqn.~\ref{eqFD2} and Eqn.~\ref{eqFD3}, we obtain the bounded difference condition for the multivariate risk functional:
\begin{equation}
    |F(D_1, \cdots, D_k, \cdots, D_m) - F(D_1, \cdots, D_k^{\prime}, \cdots, D_m)|\leq \frac{2\sqrt{2}}{m\cdot\mathrm{Inv}(\mathbb{Z})\cdot\sqrt{K(\mathbb{Z})}}
\end{equation}
\end{proof}

\section{Proof of Theorem~\ref{thm:main-bound}}\label{proof3}
\begin{proof}
We start by adding and subtracting the per-domain expected risks, then apply the triangle inequality:
    \begin{equation}
        \begin{split}
            \sup_{h \in \mathcal{H}}\mathrm{Err}_D(h) &= \sup_{h \in \mathcal{H}} \mathcal{E}_{\mathcal{P}^C}(h) - \frac{1}{m}\sum_{i=1}^m \hat{\mathcal{E}}_{S_{i}}(h) \\
            &= \sup_{h \in \mathcal{H}}E_{\mathcal{P}^C}\Big[E_D[\ell(h(x),y)]\Big]-\frac{1}{m}\sum_{i=1}^m\frac{1}{n}\sum_{j=1}^n\ell(h(x_{i,j}), y_{i, j})\\&\leq \underbrace{\sup_{h \in \mathcal{H}}E_{\mathcal{P}^C}\Big[E_D[\ell(h(x),y)]\Big]- \frac{1}{m}\sum_{i=1}^mE_{D_i}[\ell(h(x),y)]}_{\text{Term}_A}\\&\quad+\underbrace{\sup_{h \in \mathcal{H}}\frac{1}{m}\sum_{i=1}^m E_{D_i}[\ell(h(x),y)]-\frac{1}{m}\sum_{i=1}^m\frac{1}{n}\sum_{j=1}^n\ell(h(x_{i,j}), y_{i, j})}_{\text{Term}_B}
        \end{split}
    \end{equation}

    (1) We first handle Expression $\text{Term}_B$:
    \begin{equation}\label{eqB1}
        \begin{split}
            \text{Term}_B&=\sup_{h \in \mathcal{H}}\frac{1}{m}\sum_{i=1}^m E_{D_i}[\ell(h(x),y)]-\frac{1}{m}\sum_{i=1}^m\frac{1}{n}\sum_{j=1}^n\ell(h(x_{i,j}), y_{i, j})\\&\leq \frac{1}{m}\sum_{i=1}^m \sup_{h \in \mathcal{H}}\Big(E_{D_i}[\ell(h(x),y)]-\frac{1}{n}\sum_{j=1}^n\ell(h(x_{i,j}), y_{i, j})\Big)\\&=\frac{1}{m}\sum_{i=1}^m \sup_{h \in \mathcal{H}}\mathcal{E}_{D_i}(h)-\mathcal{\hat{E}}_{S_i}(h)\\&\leq \frac{2}{m}\sum_{i=1}^m \mathcal{\hat{R}}^i_n(\ell \circ \mathcal{H})+3\sqrt{\frac{\ln{(2/\delta)}}{2n}}\\&\leq \frac{2q}{m}\sum_{i=1}^m \mathcal{\hat{R}}^i_n(\mathcal{H})+3\sqrt{\frac{\ln{(2/\delta)}}{2n}}
        \end{split}
    \end{equation}
    The above inequality holds with probability at least $1-\delta$.

    (2) We next handle Expression $\text{Term}_A$.
    \begin{equation}
        \begin{split}
            \text{Term}_A&=\sup_{h \in \mathcal{H}}E_{\mathcal{P}^C}\Big[E_D[\ell(h(x),y)]\Big]- \frac{1}{m}\sum_{i=1}^mE_{D_i}[\ell(h(x),y)]\\&=\sup_{h \in \mathcal{H}}E_{\mathcal{P}^C}[\phi_{\ell, h}(D)-\frac{1}{m}\sum_{i=1}^m\phi_{\ell, h}(D_i)]\\&=\sup_{\phi \in \Phi_{\ell,\mathcal{H}}^C}E_{\mathcal{P}^C}[\phi(D)-\frac{1}{m}\sum_{i=1}^m\phi(D_i)]\\&=F(D_1, \cdots, D_k, \cdots, D_m)
        \end{split}
    \end{equation}
    According to Corollary~\ref{cor:concentration}, we have:
\begin{equation}
    \int_{\Omega^m} {\rm d}(\mathcal{P}^C)^m  \leq 2\exp\left(-\frac{mt^2\cdot\mathrm{Inv}(\mathbb{Z})^2\cdot K(\mathbb{Z})}{4}\right),
\end{equation}
    where 
    \begin{equation}
        \begin{split}
    \Omega^m &=\big\{(D_1, \cdots, D_k, \cdots, D_m) \in (\mathbb{D}^C)^m:\\& \quad\quad F(D_1, \cdots, D_k, \cdots, D_m) - E_{(\mathcal{P}^c)^m}[F(D_1, \cdots, D_k, \cdots, D_m)]\geq t\big\},
\end{split}
    \end{equation}
    Let $\delta=2\exp\left(-\frac{mt^2\cdot\mathrm{Inv}(\mathbb{Z})^2\cdot K(\mathbb{Z})}{4}\right)$, then we have:
    \begin{equation}
        \int_{\Omega^m_c} {\rm d}(\mathcal{P}^C)^m  \geq 1-\delta,
    \end{equation}
    where
    \begin{equation}
    \resizebox{0.98\linewidth}{!}{$
    \begin{split}
        \Omega^m_c=\left\{(D_1, \cdots, D_k, \cdots, D_m) \in (\mathbb{D}^C)^m:F(D_1, \cdots, D_k, \cdots, D_m) - E_{(\mathcal{P}^C)^m}[F(D_1, \cdots, D_k, \cdots, D_m)]\leq 2\sqrt{\frac{\ln(2/\delta)}{m\cdot\mathrm{Inv}(\mathbb{Z})^2\cdot K(\mathbb{Z})}}\right\}.
    \end{split}$}
    \end{equation}
    i.e. the following inequality holds with probability at least $1-\delta$:
    \begin{equation}\label{Eq_A}
        \text{Term}_A-\underbrace{E_{(\mathcal{P}^C)^m}\Big[\sup_{h \in \mathcal{H}}E_{\mathcal{P}^C}\big[E_D[\ell(h(x),y)]\big]- \frac{1}{m}\sum_{i=1}^mE_{D_i}[\ell(h(x),y)] \Big]}_{\text{Term}_C} \leq 2\sqrt{\frac{\ln(2/\delta)}{m\cdot\mathrm{Inv}(\mathbb{Z})^2\cdot K(\mathbb{Z})}}.
    \end{equation}
    Next, we handle Equation $\text{Term}_C$.
    \begin{equation}\label{Eq_R1}
        \begin{split}
            \text{Term}_C&=E_{D \sim (\mathcal{P}^C)^m}\Big[\sup_{h \in \mathcal{H}}E_{\mathcal{P}^C}\big[E_D[\ell(h(x),y)]\big]- \frac{1}{m}\sum_{i=1}^mE_{D_i}[\ell(h(x),y)] \Big]\\&=E_{D \sim (\mathcal{P}^C)^m}\Big[\sup_{h \in \mathcal{H}}E_{\mathcal{P}^C}[\phi_{\ell, h}(D)]-\frac{1}{m}\sum_{i=1}^m\phi_{\ell, h}(D_i)\Big]\\&=E_{D \sim (\mathcal{P}^C)^m}\Big[\sup_{\phi \in \Phi^C_{\ell, \mathcal{H}}}E_{\mathcal{P}^C}[\phi]-\hat{E}_D[\phi]\Big]\\&=E_{D \sim (\mathcal{P}^C)^m}\Big[\sup_{\phi \in \Phi^C_{\ell, \mathcal{H}}}E_{D^{\prime} \sim (\mathcal{P}^C)^m}\big[\hat{E}_{D^{\prime}}[\phi]-\hat{E}_D[\phi]\big]\Big]\\&\leq E_{D \sim (\mathcal{P}^C)^m}E_{D^{\prime} \sim (\mathcal{P}^C)^m}\Big[\sup_{\phi \in \Phi^C_{\ell, \mathcal{H}}}\hat{E}_{D^{\prime}}[\phi]-\hat{E}_D[\phi]\Big]\\&\leq E_{D \sim (\mathcal{P}^C)^m}E_{D^{\prime} \sim (\mathcal{P}^C)^m}\Big[\sup_{\phi \in \Phi^C_{\ell, \mathcal{H}}}\frac{1}{m}\sum_{i=1}^m\phi(D_i^{\prime})-\phi(D_i)\Big]\\& = E_{\sigma}E_{D \sim (\mathcal{P}^C)^m}E_{D^{\prime} \sim (\mathcal{P}^C)^m}\Big[\sup_{\phi \in \Phi^C_{\ell, \mathcal{H}}}\frac{1}{m}\sum_{i=1}^m \sigma_i \big(\phi(D_i^{\prime})-\phi(D_i)\big)\Big]\\&\leq E_{\sigma}E_{D \sim (\mathcal{P}^C)^m}\Big[\sup_{\phi \in \Phi^C_{\ell, \mathcal{H}}}\frac{1}{m}\sum_{i=1}^m \sigma_i \phi(D_i)\Big] + E_{\sigma}E_{D^{\prime} \sim (\mathcal{P}^C)^m}\Big[\sup_{\phi \in \Phi^C_{\ell, \mathcal{H}}}\frac{1}{m}\sum_{i=1}^m -\sigma_i \phi(D_i^{\prime})\Big]\\&=2E_{\sigma}E_{D \sim (\mathcal{P}^C)^m}\Big[\sup_{\phi \in \Phi^C_{\ell, \mathcal{H}}}\frac{1}{m}\sum_{i=1}^m \sigma_i \phi(D_i)\Big]\\&=2E_{D \sim (\mathcal{P}^C)^m}[\hat{\mathcal{R}}_{D_1, \cdots, D_m}(\Phi^C_{\ell, \mathcal{H}})]\\&=2E_{D \sim (\mathcal{P}^C)^m}[\hat{\mathcal{R}}_{D_1, \cdots, D_m}(\Phi_{\ell, \mathcal{H}})]
        \end{split}
    \end{equation}
    Let $\hat{\mathcal{R}}_{D_1, \cdots, D_k^{\prime}, \cdots, D_m}(\Phi_{\ell, \mathcal{H}})$ be the Rademacher complexity after replacing the distribution of the k-th source domain. Analogously to the multivariate risk functionals $F$, we simplify the derivation in the proof of Theorem~\ref{thm:bounded-difference} to obtain:
    \begin{equation}
        \begin{split}
            \hat{\mathcal{R}}_{D_1, \cdots, D_k, \cdots, D_m}(\Phi_{\ell, \mathcal{H}})-\hat{\mathcal{R}}_{D_1, \cdots, D_k^{\prime}, \cdots, D_m}(\Phi_{\ell, \mathcal{H}})& \leq \sup_{h \in \mathcal{H}}\frac{|\phi_{\ell, h}(D_k)-\phi_{\ell, h}(D_k^{\prime})|}{m}\\& \leq \frac{1}{m}\Big(\sqrt{\frac{2H_{\infty, \nu}(D_k)}{K(\mathbb{Z})}}+\sqrt{\frac{2H_{\infty, \nu}(D_k^{\prime})}{K(\mathbb{Z})}}\Big)\\& \leq \frac{2\sqrt{2}}{m\cdot\mathrm{Inv}(\mathbb{Z})\cdot\sqrt{K(\mathbb{Z})}}
        \end{split}
    \end{equation}
    Similarly, we have:
    \begin{equation}
        \hat{\mathcal{R}}_{D_1, \cdots, D_k^{\prime}, \cdots, D_m}(\Phi_{\ell, \mathcal{H}}) - \hat{\mathcal{R}}_{D_1, \cdots, D_k, \cdots, D_m}(\Phi_{\ell, \mathcal{H}}) \leq \frac{2\sqrt{2}}{m\cdot\mathrm{Inv}(\mathbb{Z})\cdot\sqrt{K(\mathbb{Z})}}
    \end{equation}
    so, 
    \begin{equation}
        \Big|\hat{\mathcal{R}}_{D_1, \cdots, D_k, \cdots, D_m}(\Phi_{\ell, \mathcal{H}})-\hat{\mathcal{R}}_{D_1, \cdots, D_k^{\prime}, \cdots, D_m}(\Phi_{\ell, \mathcal{H}})\Big| \leq \frac{2\sqrt{2}}{m\cdot\mathrm{Inv}(\mathbb{Z})\cdot\sqrt{K(\mathbb{Z})}}
    \end{equation}
    so $\hat{\mathcal{R}}_{D_1, \cdots, D_m}$ also satisfies Theorem~\ref{thm:bounded-difference}, and applying Corollary~\ref{cor:concentration}, the following inequality holds with probability at least $1-\delta$:
    \begin{equation}\label{Eq_R2}
    \begin{split}
        E_{D \sim (\mathcal{P}^C)^m}\big[\hat{\mathcal{R}}_{D_1, \cdots, D_m}(\Phi_{\ell, \mathcal{H}})\big] &\leq \hat{\mathcal{R}}_{D_1, \cdots, D_m}(\Phi_{\ell, \mathcal{H}})+2\sqrt{\frac{\ln(2/\delta)}{m\cdot\mathrm{Inv}(\mathbb{Z})^2\cdot K(\mathbb{Z})}}.
    \end{split} 
    \end{equation}

    Combining Eqn.~\ref{eqB1}, Eqn.~\ref{Eq_A}, Eqn.~\ref{Eq_R1}, and Eqn.~\ref{Eq_R2}, we obtain the final result:
    \begin{equation}\small
        \sup_{h \in \mathcal{H}}Err_D(h) \leq 2\hat{\mathcal{R}}_{D_1, \cdots, D_m}(\Phi_{\ell, \mathcal{H}}) + \frac{2q}{m}\sum_{i=1}^m \mathcal{\hat{R}}^i_n(\mathcal{H}) + \mathcal{O}\Big(\sqrt{\frac{\ln(2/\delta)}{n}}\Big)+\mathcal{O}\Big( \sqrt{\frac{\ln(2/\delta)}{m\cdot\mathrm{Inv}(\mathbb{Z})^2\cdot K(\mathbb{Z})}}\Big). 
    \end{equation}
    The above inequality holds with probability at least $1-3\delta$ by a union bound over the three high-probability events used above.
\end{proof}



\section{Experimental Details}\label{exp2}
\subsection{Dataset Introduction}\label{expD}
We provide a brief overview of the datasets used in these experiments.

\textbf{Digits-DG} \citep{zhou2020deep} comprises four distinct digit datasets, namely MNIST \citep{Yann32}, MNIST-M \citep{Ganin19}, SVHN \citep{Yuval34}, and SYN \citep{Ganin19}, which exhibit substantial variations in font style, stroke color, and background. Each domain contains ten categories.

\textbf{PACS} \citep{Lid31} consists of four domains, namely Photo (1,670 images), Art Painting (2,048 images), Cartoon (2,344 images) and Sketch (3,929 images). Each domain contains seven categories. 

\textbf{Office-Home} \citep{Venkateswara36} contains four domains,which are Artistic, Clipart, Product and Real-World. Each domain has 65 classes, which are related to office and home objects. There are around 15,500 images in total. The domain variations mainly take place in background, view point and image style.

\subsection{Implementation Details in Section~\ref{exp4.1}}\label{app:impl_details 1}
\paragraph{von Mises--Fisher distribution.}
The von Mises--Fisher (vMF) distribution provides a natural framework for modeling data distributed on a hypersphere \citep{ding115, xu114}. It is
characterized by a mean direction $\boldsymbol{\mu} \in \mathcal{S}^{m-1}$ and a
concentration parameter $\kappa \ge 0$, with support on the unit hypersphere
$\mathcal{S}^{m-1} \subset \mathbb{R}^m$. Sampling from a vMF distribution yields
latent codes that inherently reside on a hyperspherical manifold. The
concentration parameter $\kappa$ governs the tightness of samples around
$\boldsymbol{\mu}$; in the limiting case where $\kappa = 0$, the vMF distribution
reduces to the uniform distribution over the sphere, thereby achieving maximum
coverage. 
\paragraph{Encoder and mixture-of-spheres construction.}
An encoder network maps each input image $x$ to the parameters
$(\boldsymbol{\mu}, \kappa)$ of a vMF distribution. Specifically, the encoder is built upon a pretrained ResNet-18 convolutional backbone, followed by three
independent linear heads that respectively predict the vMF parameters $(\boldsymbol{\mu}, \kappa)$ and the mixture-assignment logits. We then draw a
sample $\mathbf{z}_{\text{unit}} \sim \mathrm{vMF}(\boldsymbol{\mu}, \kappa)$, which lies on the unit hypersphere.

While the above approach is theoretically sound, we empirically observed that a
single hypersphere with a small radius (high curvature) tends to cause training
difficulties. This is due to a decrease in representational capacity, as the data
points become densely packed together. To overcome this limitation while
retaining the benefits of curvature control, we introduce a mixture of multiple
hyperspheres with learnable centers. Concretely, instead of using a single
spherical manifold, we construct a latent space consisting of a mixture of
$N_{\text{sph}}$ hyperspheres. These hyperspheres share an equal radius $R$, but
have distinct centers. The center of the $i$-th sphere is represented by a
learnable vector $\mathbf{c}_i \in \mathbb{R}^{m}$, where
$i \in [1, N_{\text{sph}}]$. We stochastically sample a latent variable via
standard mixture density sampling. Specifically, we first determine the sphere
index using the mixture weights $\boldsymbol{\pi}$. These weights are derived by
computing the logits $\boldsymbol{\pi}_{\text{logits}}$ of an
$N_{\text{sph}}$-categorical distribution over the spheres, followed by a
Gumbel--Softmax operation to obtain a differentiable assignment vector
$\boldsymbol{\pi} \in \triangle^{N_{\text{sph}}}$, i.e., an $N_{\text{sph}}$-dimensional simplex, such that $\sum_{i=1}^{N_{\text{sph}}} \pi_i = 1$:
\begin{equation}
\boldsymbol{\pi} = \text{Gumbel-Softmax}(\boldsymbol{\pi}_{\text{logits}}, \tau),
\end{equation}
where $\tau$ is a temperature parameter (annealed during training). The final
latent representation is a convex combination of the sphere-specific points:
\begin{equation}
\mathbf{z} = R \cdot \mathbf{z}_{\text{unit}}
+ \sum_{i=1}^{N_{\text{sph}}} \pi_i \mathbf{c}_i.
\end{equation}
Thus, $\mathbf{z}$ lies on the union of $N_{\text{sph}}$ spheres of radius $R$,
each centered at $\mathbf{c}_i$, and the assignment $\boldsymbol{\pi}$ controls
which sphere contributes most to the representation. The learnable centers
$\mathbf{c}_i$ are initialized on a sphere of radius $3$ and are regularized to
stay well-separated, ensuring that the composite geometry remains expressive
while preserving curvature properties. Importantly, the curvature of each
individual sphere is still $\mathcal{C} = 1/R^2$, so controlling $R$ directly
influences the geometric regularization of the representation.
\paragraph{Classifier architecture.}
The scaled latent code $\mathbf{z}$ is then passed to a classifier that predicts
the class label. For PACS and Digits-DG, the classifier is a single linear layer,
the latent space dimensions are set to $m = 5, 10$, with the number of
hypersphere centers $N_{\text{sph}}$ correspondingly chosen as $\{7, 15\}$. For
Office-Home, which has 65 categories, we use a two-layer network with 64 hidden
units, ReLU activation, and a dropout rate of 0.3. The latent space dimension is
set to $m = 20$, and the number of hypersphere centers is set to
$N_{\text{sph}} = 75$. The entire pipeline is trained end-to-end by minimizing a
supervised classification loss; no reconstruction term or KL divergence is
required, as our focus is solely on the discriminative power of the curved latent
space. The radius $R$ governs the curvature of the embedding space and is treated
as a fixed hyperparameter.
\paragraph{Training procedure.}
We summarize the experimental procedure as Algorithm~\ref{alg:training}.
\begin{algorithm}[t]
\caption{Training procedure for the proposed model}
\label{alg:training}
\begin{algorithmic}[1]
\State \textbf{Initialize} all parameters randomly:
\State \hspace{0.5em} encoder backbone,
\State \hspace{0.5em} learnable sphere centers $\{\mathbf{c}_i\}_{i=1}^{N_{\text{sph}}}$,
\State \hspace{0.5em} classifier $f_{\text{cls}}$.
\Repeat
    \State Sample a batch $(x, y)$ from the union of source domains.
    \State Encode to obtain $(\boldsymbol{\mu}, \kappa) = \big(\text{Encoder}_1(x), \text{Encoder}_2(x)\big)$ with $\|\boldsymbol{\mu}\|_2 = 1$.
    \State Obtain mixture assignment logits $\boldsymbol{\pi}_{\text{logits}} = \text{Encoder}_3(x)$.
    \State Compute $\boldsymbol{\pi} = \text{GumbelSoftmax}(\boldsymbol{\pi}_{\text{logits}}, \tau)$.
    \State Sample $u \sim \mathrm{vMF}(\boldsymbol{\mu}, \kappa)$.
    \State Form the representation: $z = R \cdot u + \sum_{i} \pi_i \mathbf{c}_i$.
    \State Predict $\hat{y} = f_{\text{cls}}(z)$ and compute the cross-entropy loss.
    \State Update all learnable parameters.
\Until{convergence}
\end{algorithmic}
\end{algorithm}
\paragraph{Radius and curvature settings.}
We configure the radius settings separately for different datasets in our
experiments. Specifically, for PACS and Digits-DG, we choose radii
$R \in \{5, 2, 1\}$, corresponding to curvatures
$\mathcal{C} = 0.04,\ 0.25,\ 1$, respectively. For Office-Home, which contains
65 categories and is more challenging to train, we adopt larger radii
$R \in \{9, 7, 5\}$, yielding curvatures
$\mathcal{C} \approx 0.012,\ 0.020,\ 0.040$. This choice is motivated by the fact
that Office-Home has a significantly larger number of classes than the other two
datasets; a small radius in the latent space would lead to excessive
representation collapse on the hypersphere, making it difficult to preserve
sufficient discriminative information and thereby hindering training convergence
and limiting accuracy improvement. Therefore, we employ larger radii on
Office-Home to provide greater representational capacity, while still using the
same mixture-of-spheres architecture across all datasets. This enables the model
to adaptively assign different data patterns to different centers, effectively
mitigating the training difficulty associated with a single sphere at small radii
while continuing to benefit from the curvature-driven generalization guarantee.
\paragraph{Discussion of the Photo anomaly.}
In most cases, increasing the latent space curvature (i.e., decreasing the
hyperspherical radius) leads to a smaller domain generalization error. A notable
exception is the case where Photo is used as the target domain in PACS: the
domain generalization error remains consistently low across all curvature values,
and no clear relationship between curvature and generalization performance is
evident. We hypothesize that this anomaly arises because the target Photo domain
may be particularly similar to one of the source domains (e.g., Art Painting,
Cartoon, or Sketch). In such a scenario, the distribution shift between the
source domains and the target is minimal, allowing the model to achieve strong
generalization even without curvature adjustments. Consequently, the beneficial
effect of a more curved latent space becomes less pronounced, as the inherent
simplicity of the transfer task overshadows any gains from curvature.

\subsection{Implementation Details in Section~\ref{exp4.2}}\label{app:impl_details 2}
\paragraph{Source domain configurations.}
For the three benchmark datasets (PACS, Office-Home, and Digits-DG), we consider
the following source domain configurations:
\begin{enumerate}
\item \textbf{Single source domain:} randomly select one domain as the source
and evaluate its generalization performance on each of the remaining three
domains.
\item \textbf{Two source domains:} randomly select a pair of domains as the
source set and evaluate on the other two domains.
\item \textbf{Three source domains:} use all three domains as sources (standard
multi-source setting) and evaluate on the remaining target domain.
\item \textbf{Six source domains via MixStyle:} apply the MixStyle
\citep{Zhou52} strategy to mix the feature statistics of source domains,
generating three mixed domains (each obtained by mixing a different pair of
source domains), resulting in a total of six source domains.
\end{enumerate}

\paragraph{MixStyle.}
MixStyle \citep{Zhou52} is a method that generates new domain samples by
computing style statistics and mixing the style statistics of samples from
different source domains.

\paragraph{Experimental setup.}
For each dataset and each target domain, we train models under the four source
configurations described above, using a pretrained ResNet-18 as the backbone,
where the hypothesis class corresponds to the classification head of the
network. The domain generalization error is defined as the test error on the
target domain minus the average training error on the source domains. The network
is trained for 50 epochs, and the error is averaged over the final five epochs to
reduce random fluctuations.

\paragraph{Visualization.}
We visualize the results as bar charts, where each bar corresponds to a specific
triple of (dataset, target domain, source configuration). For each dataset, we
show four subfigures corresponding to the four target domains, resulting in
twelve bar charts in total. In each chart, the horizontal axis denotes the
different source configurations (single source, two sources, three sources, six
sources via MixStyle), and the vertical axis represents the domain generalization
error. Error bars indicate the standard deviation across three runs.

\paragraph{Discussion of exceptions.}
As shown in Figure~\ref{fig12}, the results are consistent with our theoretical bound in
most cases: increasing the number of source domains generally reduces the domain
generalization error. However, when Photo serves as the target domain on PACS,
the results exhibit notable random fluctuations. Moreover, on Office-Home, when
Clipart is used as the target domain, the influence of the number of source
domains on the generalization error is less pronounced.

We conjecture that this may be due to the inherent characteristics of these
specific domains. For PACS with Photo as the target, the Photo domain may be
relatively similar to the source domains (e.g., Art Painting, Cartoon, Sketch),
making the model less sensitive to the number of source domains and leading to
unstable estimates. In the case of Office-Home with Clipart as the target,
Clipart may present a unique distribution that is less well-covered by the source
domains, causing the model to rely more heavily on domain-invariant features and
thereby diminishing the observable effect of varying source domain quantity.

\subsection{Experimental Limitations}\label{Exp-Lim}
A limitation of our experimental design is that using a single hypersphere can lead to underfitting, and although the mixture-of-hyperspheres construction is introduced to mitigate this issue, it may not eliminate it entirely. In practice, we do observe variations in training accuracy across different radii. Consequently, our empirical validation of the relationship between curvature and generalization error relies on observing changes in the generalization gap, which introduces a degree of experimental imprecision. Ideally, one would isolate the curvature effect while holding training accuracy exactly fixed; however, this is difficult to achieve in practice because changing the radius directly affects representation capacity and optimization. We therefore report the generalization gap as supporting evidence rather than as a decisive causal validation, and we explicitly acknowledge this limitation in the interpretation of our experimental results.

\end{document}